\documentclass[12pt]{article}

\usepackage[utf8]{inputenc}
\usepackage[fleqn]{amsmath}
\usepackage{amssymb}
\usepackage{mathtools}
\usepackage{times}
\usepackage{graphicx}
\usepackage[export]{adjustbox}
\usepackage{color}
\usepackage{multirow}
\usepackage{apacite}
\usepackage{rotating}
\usepackage{bbm}
\usepackage{latexsym}
\usepackage[separate-uncertainty=true]{siunitx}
\usepackage{acronym}
\usepackage{natbib}
\usepackage{epsfig}
\usepackage{subfigure}
\usepackage{svn-multi}
\numberwithin{equation}{section}
\usepackage{amssymb}
\usepackage[hidelinks]{hyperref}

\acrodef{STDP}{spike-timing-dependent plasticity}
\acrodef{PSP}{postsynaptic potential}
\acrodef{SRM}[SRM\textsubscript{0}]{spike response model}
\acrodef{SNN}{spiking neural network}
\acrodef{LIF}{leaky integrate-and-fire}
\acrodef{ReLU}{rectified linear unit}
\acrodef{SEM}{standard error of the mean}
\acrodef{ANN}{artificial neural network}
\acrodef{XOR}{exclusive-or}

\newcommand{\captionfonts}{\normalsize}

\makeatletter  
\long\def\@makecaption#1#2{%
  \vskip\abovecaptionskip
  \sbox\@tempboxa{{\captionfonts #1: #2}}%
  \ifdim \wd\@tempboxa >\hsize
    {\captionfonts #1: #2\par}
  \else
    \hbox to\hsize{\hfil\box\@tempboxa\hfil}%
  \fi
  \vskip\belowcaptionskip}
\makeatother   

\begin{document}
\hspace{13.9cm}

\ \vspace{20mm}\\

\noindent
{\LARGE \bf Supervised Learning with First-to-Spike \\Decoding in Multilayer Spiking Neural Networks}\\

\ \\
{\bf \large Brian Gardner$^{1, \ast}$ \& Andr\'{e} Gr\"{u}ning$^{2}$}\\
$^{1}$Department of Computer Science\\
University of Surrey\\
Guildford\\
United Kingdom\\
\ \\
$^{2}$Faculty of Electrical Engineering and Computer Science\\
University of Applied Sciences Stralsund\\
Germany\\
\ \\
$^{\ast}$Corresponding author.\\E-mail: b.gardner@surrey.ac.uk

\thispagestyle{empty}

\section*{Abstract}

Experimental studies support the notion of spike-based neuronal information processing in the brain, with neural circuits exhibiting a wide range of temporally-based coding strategies to rapidly and efficiently represent sensory stimuli. Accordingly, it would be desirable to apply spike-based computation to tackling real-world challenges, and in particular transferring such theory to neuromorphic systems for low-power embedded applications.
Motivated by this, we propose a new supervised learning method that can train multilayer spiking neural networks to solve classification problems based on a rapid, first-to-spike decoding strategy. The proposed learning rule supports multiple spikes fired by stochastic hidden neurons, and yet is stable by relying on first-spike responses generated by a deterministic output layer. In addition to this, we also explore several distinct, spike-based encoding strategies in order to form compact representations of presented input data.
We demonstrate the classification performance of the learning rule as applied to several benchmark datasets, including MNIST. The learning rule is capable of generalising from the data, and is successful even when used with constrained network architectures containing few input and hidden layer neurons. Furthermore, we highlight a novel encoding strategy, termed `scanline encoding', that can transform image data into compact spatiotemporal patterns for subsequent network processing. Designing constrained, but optimised, network structures and performing input dimensionality reduction has strong implications for neuromorphic applications.

\section{Introduction}

Neurons constitute complex biological circuits, and work to convey information via rapid, spike-based signalling. These neural circuits interconnect with one another, forming the basis of large scale networks in the brain, and are often organised as consecutive processing layers operating at increasing levels of abstraction. For example, within the visual system, information regarding object features can be temporally encoded as spikes in little over just 10 ms, and its identity determined through feedforward processing pathways within 200 ms of pattern onset \citep{Kiani2005,Gollisch2008,Hung2005}.
In terms of forming such representations, the adaptation of synaptic connections between neurons is hypothesised to underlie the learning process: principally based on correlated neuronal activity patterns and regulatory, homeostatic plasticity mechanisms \citep{Morrison2008}. In particular, a Hebbian-like learning scheme, termed \ac{STDP}, is considered to play a prominent role \citep{Gerstner2002}, whereby the strength of a synaptic connection is modified according to the relative timing difference between paired pre- and postsynaptic firing events \citep{Bi1998}. Drawing on these principles, theoretical work has sought to model goal-directed learning in the brain using \acp{SNN}: typically incorporating concepts from machine learning such as supervised and reinforcement learning for this purpose \citep{Gruning2014}. Despite progress in this respect, a more comprehensive theoretical description of learning that also aims to more fully exploit the rapidity and precision of spike-based temporal coding is still largely lacking; consequently, finding real-world applications for spike-based learning and computing techniques, including their transfer to neuromorphic platforms, remains an open issue.

For the most part, theoretical studies into spike-based learning in neural networks have been devised based on the dynamics of the \ac{LIF} neuron model, typically when formalised as the simplified \acf{SRM} \citep{Gerstner2002}, owing to its convenient trade-off between analytical tractability and model realism in run simulations. Additionally, the application of gradient descent in order to minimise the value of some predefined cost function, as taken for supervised learning, is a useful starting point in order to obtain weight update rules for \acp{SNN} \citep{Gutig2014}. A common learning objective has been to train neurons to precisely fire at one or more prescribed target firing times; to this end, the cost function of an \ac{SNN} is usually defined in terms of the separation between target and actual firing times with respect to one or more of its readout neurons. Hence, by applying gradient descent, weight update rules for the network can be derived, and accordingly implemented during training in order to support neuronal firing at these target timings \citep{Bohte2002a,Florian2012,Sporea2013,Gardner2015,Zenke2018}.
Furthermore, some supervised approaches have incorporated a trained neuron's subthreshold voltage into the network cost function: for example to support a more efficient mode of operation in addition to learning target firing times \citep{Albers2016}. In a similar effort, the minimum distance between a neuron's voltage and its firing threshold, as measured over some predetermined observation period, has been selected as the point at which a neuron should be driven to fire, in order to provide a highly efficient spike-based classifier rule \citep{Gutig2006,Urbanczik2009}.
Aside from setting up an initial cost function, some studies have instead taken a statistical approach to learning; for instance, based on a maximum-likelihood principle that works to maximise the likelihood of an \ac{SNN} generating desired target firing times \citep{Pfister2006, Gardner2016}, or similarly by minimising an upper bound on the KL-divergence between the actual and target firing distribution for more complex, recurrent \ac{SNN} architectures \citep{Brea2013}. A variational, online learning rule for training recurrent \acp{SNN} has also recently been proposed in \cite{Jang2020}; a detailed review of probabilistic learning methods can be found in \cite{Jang2019}. Otherwise, the procedure used to learn target firing times may be mapped from Perceptron-like learning for high precision \citep{Memmesheimer2014}, or more heuristically by modifying weights according to a spike-based adaptation of the Widrow-Hoff rule \citep{Ponulak2010,Mohemmed2012,Yu2013}.

A large part of the studies described so far have been concerned with training \acp{SNN} to learn target output firing times, which in most cases tend to be arbitrarily selected; this usually follows from focusing on a proof-of-concept of a derived learning procedure, rather than measuring its technical performance on benchmark datasets. Moreover, biological plausibility is a common concern with spike-based learning approaches, which can place further constraints on a model and detract from its performance. Although there is likely to be strong potential in utilising spike-based computation for data classification purposes, it remains unclear which temporal coding strategy is best suited for this purpose. For instance, learning multiple, temporally-precise sequences of spikes in order to categorise input patterns into different classes might inadvertently lead to model overfitting, and hinder generalisation to previously unseen samples; this is more likely to be an issue with high precision rules, for example the E-learning variant of the Chronotron \citep{Florian2012} or the HTP algorithm \citep{Memmesheimer2014}. Therefore, a preferable coding strategy for a spike-based classifier might instead rely on selecting output responses which place the least constraint on the trained parameters: for example as demonstrated by the single-layer Tempotron rule \citep{Gutig2006}.

Although the minimally-constrained Tempotron has proven capable of high performance with respect to certain problem domains, such as vocabulary recognition \citep{Gutig2009}, there may arise limitations in terms of its flexibility as applied to increasingly challenging datasets such as MNIST, for which networks containing hidden neurons are indicated for its solution; interestingly, however, recently submitted work has addressed this issue by implementing a Tempotron-inspired cost function combined with a multilayer learning procedure, and to good effect \citep{Zenke2020}. Despite such progress, there still exist comparatively few learning rules for multilayer \acp{SNN} compared with single-layer ones, owing to the complexity in solving ill-defined gradients of hidden layer spike trains when applying the technique of backpropagation. A number of approaches have relied on approximating such gradients: for example by taking a linear approximation of a neuron's response close to its firing threshold \citep{Bohte2002a}, estimating a spike train by its underlying firing density \citep{Sporea2013} or using a surrogate gradient to substitute a neuron's spike-gradient with an analytically tractable one \citep{Zenke2018, Neftci2019}. Furthermore, some studies have taken a statistical approach which instead consider the likelihood of a neuron's firing response, as applied to feedforward \citep{Gardner2015} and recurrent \citep{Brea2013,Rezende2014} network structures. In these cases, however, the networks have been constrained to learning predefined, target firing patterns, with less of a focus on utilising efficient temporal encoding and decoding strategies for data classification purposes. Of the studies which have focused on applying multilayer or deep \acp{SNN} to more challenging datasets, some have demonstrated that training rate-based \acp{ANN} and transferring the learned weights to similarly designed \acp{SNN} for test inference can provide performance competitive with state-of-the-art systems \citep{Connor2013,Diehl2015}. A main limitation of this approach, however, is that these equivalent \acp{ANN} must be trained offline before being mapped to an online system, making this technique somewhat restrictive in terms of its application to adaptive learning tasks. A further study exploring deep \ac{SNN} architectures considered a scheme which involved low-pass filtering spike events in order to establish smooth gradients for backpropagation, although this came with the caveat of introducing auxiliary variables which needed to be computed separately \citep{Lee2016}. Other studies have arrived at alternative solutions by approximating simulated \ac{LIF} neurons as \acp{ReLU} \citep{Tavanaei2019}, or instead simulating non-leaky integrate-and-fire neurons for analytical tractability \citep{Mostafa2017}, thereby establishing closed-form expressions for the weight updates. These methods have resulted in competitive performance on the MNIST dataset, and the first-to-spike decoding method implemented by \cite{Mostafa2017}, which classifies data samples according to which output neuron is the first to respond with a spike, has proven to be particularly rapid at forming predictions. Interestingly, the recent work of \cite{Bagheri2018}, which examined training probabilistic, single-layer \acp{SNN} as applied to MNIST, has also indicated at the merits of utilising a rapid, first-to-spike decoding scheme.

In this article, we introduce a new supervised learning algorithm to train multilayer \acp{SNN} for data classification purposes, based on a first-to-spike decoding strategy. This algorithm extends on our previous MultilayerSpiker rule described in \cite{Gardner2015}, by redefining the network's objective function as a cost over first spike arrival times in the output layer, and instead implementing deterministic output neurons for more robust network responses. We test our new first-to-spike multilayer classifier rule on several benchmark classification tasks, including the ubiquitous MNIST dataset of handwritten digits, in order to provide an indication of its technical capability. Additionally, we explore several different spike-based encoding strategies to efficiently represent the input data, including one novel technique that can transform visual patterns into compact spatio-temporal patterns via `scanline encoding'. We determine that such an encoding strategy holds strong potential when applied to constrained network architectures, as might exist with a neuromorphic hardware platform. In the next section we start our analysis by describing the specifics of our first-to-spike neural classifier model.

\section{Methods}

\subsection{Neuron Model} \label{subsec:nrn_model}

We consider the simplified \ac{SRM}, as defined in \cite{Gerstner2002}, to describe the dynamics of a postsynaptic neuron's membrane potential with time $t$:
\begin{equation} \label{eq:SRM}
u_i(t) := \sum_{j \in \Gamma_i} w_{ij} \left( \epsilon \ast S_j \right)(t) + \left( \kappa \ast S_i \right)(t)\;,
\end{equation}
where the neuron is indexed $i$, and its membrane potential is measured with respect to an arbitrary resting potential. The first term on the RHS of the above equation describes a weighted sum over the neuron's received presynaptic spikes, where $\Gamma_i$ denotes the set of direct neural predecessors of neuron $i$, or its presynaptic neurons, and the parameter $w_{ij}$ refers to the synaptic weight projecting from presynaptic neuron $j$. The term $\left( \epsilon \ast S_j \right)(t) \equiv \int_0^t \epsilon(s) S_j(t - s) \mathrm{d}s$ refers to a convolution of the \ac{PSP} kernel $\epsilon$ and the $j^\mathrm{th}$ presynaptic spike train $S_j$, where a spike train is formalised as a sum of dirac-delta functions, $S_j(t) = \sum_f \delta_D (t - t_j^f)$, over a list of presynaptic firing times $\mathcal{F}_j = \{t_j^1, t_j^2, \dots\}$. The second term on the RHS of Eq~\eqref{eq:SRM} signifies the dependence of the postsynaptic neuron on its own firing history, where $\kappa$ is the reset kernel and $S_i$ is the neuron's spike train for postsynaptic firing times $\mathcal{F}_i = \{t_i^1, t_i^2, \dots\}$. A postsynaptic spike is considered to be fired at time $t_i^f$ when $u_i$ crosses the neuron's fixed firing threshold $\vartheta$ from below. The \ac{PSP} and reset kernels are, respectively, given by:
\begin{align}
\epsilon(s)
  &= \epsilon_0 \left[ \exp\left( -\frac{s}{\tau_m} \right) - \exp\left( -\frac{s}{\tau_s} \right) \right] \Theta(s) \;, \label{eq:PSP_kernel} \\
\kappa(s)
  &= \kappa_0 \exp\left(-\frac{s}{\tau_m}\right) \Theta(s) \;. \label{eq:reset_kernel}
\end{align}
With respect to Eq~\eqref{eq:PSP_kernel}, $\epsilon_0 = \SI{4}{mV}$ is a scaling constant, $\tau_m = \SI{10}{ms}$ the membrane time constant, $\tau_s = \SI{5}{ms}$ a synaptic time constant and $\Theta(s)$ the Heaviside step function. With respect to Eq~\eqref{eq:reset_kernel}, the reset strength is given by $\kappa_0 = -(\vartheta - u_r)$, where $u_r = \SI{0}{mV}$ is the value the neuron's membrane potential is reset to immediately after a postsynaptic spike is fired upon crossing the threshold $\vartheta = \SI{15}{mV}$. From Eq~\eqref{eq:SRM} it follows that the neuron's resting potential is equal to its reset value, i.e. $u_\mathrm{rest} = u_\mathrm{r} = \SI{0}{mV}$.

\subsection{Learning Rule} \label{subsec:learning_rule}

\subsubsection{Notation}

The technique of backpropagation is applied to a feedforward multilayer \ac{SNN} containing hidden layers of neurons, where the objective of the network is to perform pattern recognition on multiple input classes by learning error-minimising weights. Network layers are indexed by $l$, with $l \in \{1, 2, \dots, L - 1, L\}$, where $l = 1, L$ correspond to the input and last layers, respectively. The number of neurons in the $l^\mathrm{th}$ layer is denoted $N_l$. In our analysis, each input class corresponds to a distinct output neuron: hence, if the total number of classes is equal to $c$ then the number of output neurons is given by $N_L = c$. Using this notation, the \ac{SRM} defined by Eq~\eqref{eq:potential} is rewritten as
\begin{equation} \label{eq:potential}
u_i^l(t) := \sum_{j \in \Gamma_i^l} w_{ij}^l \left( \epsilon \ast S_j^{l-1} \right)(t) + \left( \kappa \ast S_i^l \right)(t) \;,
\end{equation}
for a postsynaptic neuron in the $l^\mathrm{th}$ layer receiving its input from previous layer neurons belonging to the set $\Gamma_i^l$. The spike train of a neuron $i$ in layer $l$ is now denoted by $S_i^l(t) = \sum_f \delta_D (t - t_i^f)$, and its associated list of firing times, $\mathcal{F}_i^l = \{t_i^1, t_i^2, \dots\}$.

\subsubsection{Cost Function}

The objective is to train a multilayer \ac{SNN} to efficiently classify input patterns based on a temporal decoding scheme. To this end, a first-to-spike code seems appropriate, since it encourages rapidity of neural processing and avoids arbitrarily constraining the network to generate spikes with specific timings. There is also experimental evidence supporting the notion of a latency code in relation to visual and neural processing pathways \citep{Hung2005,Gollisch2008}. Hence, we focus on implementing a minimally-constrained, competitive learning scheme: such that the output neuron with the earliest and strongest activation, resulting in a first-spike response, decides the class of input pattern.

Taking the above points into consideration, a suitable choice for the $i^\mathrm{th}$ output layer neuron's activation is a softmax, given by
\begin{equation} \label{eq:softmax}
a_i^L = \frac{\exp\left(-\nu \tau_i\right)}{\sum_{i'}\exp\left(-\nu \tau_{i'}\right)} \;,
\end{equation}
where $\nu$ is a scale parameter controlling the sharpness of the distribution, $i'$ indexes each output neuron, $1 \leq i' \leq c$, and $\tau_{i'}$ is the first firing time of neuron $i'$. If a neuron $i'$ fails to fire any spike, then it is assumed $\tau_{i'} \rightarrow \infty$. The set of activations can be interpreted as a conditional probability distribution over the predicted class labels. A natural choice of cost function using softmax activation is the cross-entropy, given by
\begin{equation} \label{eq:cross-entropy}
C(\mathbf{y}, \mathbf{a}^L) = -\sum_i y_i \log a_i^L \;,
\end{equation}
where $\mathbf{y} \in \mathbb{R}^c$ is a $c$-dimensional target activation vector of the network, associated with the presented input pattern, and $\mathbf{a}^L \in \mathbb{R}^{N_L}$ is the vector of output layer neuron activations. Since we are concerned with a classification problem a one-hot encoding scheme is used to describe a target vector, such that all components of $\mathbf{y}$ are set to zero except for the one corresponding to the pattern class. For example, if a dataset were comprised of three input pattern classes, then patterns belonging to the second class would be associated with $\mathbf{y} = (0, 1, 0)$. Hence, using this coding strategy, and using $y$ to denote the index of the target class label, Eq~\eqref{eq:cross-entropy} reduces to
\begin{equation} \label{eq:cross-entropy-onehot}
	C(\mathbf{y}, \mathbf{a}^L) = -\log a_y^L \;,
\end{equation}
where $a_y^L$ now denotes the activation of the single output neuron corresponding to the correct class. The above choices of cost and activation functions is inspired by the approach taken in \cite{Mostafa2017}, although here we instead consider \ac{LIF} neurons and extend our analysis to include entire spike trains generated by input and hidden layer neurons.

\subsubsection{Error Signal}

The technique of backpropagation is applied in order to train weights within the multilayer network, by minimising the cross-entropy loss defined by Eq~\eqref{eq:cross-entropy-onehot}. We begin by taking the gradient of Eq~\eqref{eq:cross-entropy-onehot} with respect to the membrane potential of a neuron $i$ in the final layer, a term which will be useful later:
\begin{equation} \label{eq:grad_cost}
\frac{\partial C(\mathbf{y}, \mathbf{a}^L)}{\partial u_i^L} = -\frac{\partial \log a_y^L}{\partial u_i^L} \;,
\end{equation}
which can be rewritten, using the chain rule, as
\begin{equation} \label{eq:grad_cost2}
\frac{\partial C(\mathbf{y}, \mathbf{a}^L)}{\partial u_i^L} = -\frac{1}{a_y^L} \frac{\partial a_y^L}{\partial u_i^L} \;.
\end{equation}
Furthermore, the gradient of the neuron's activation can be expanded using the chain rule as follows:
\begin{equation} \label{eq:grad_cost3}
\frac{\partial a_y^L}{\partial u_i^L} = \frac{\partial a_y^L}{\partial \tau_i} \frac{\tau_i}{\partial u_i^L} \;.
\end{equation}
Using Eq~\eqref{eq:softmax}, the first gradient on the RHS of the above can be solved to provide one of two cases:
\begin{equation} \label{eq:grad_cost4}
\frac{\partial a_y^L}{\partial \tau_i} = 
	\begin{cases}
		a_y^L (a_i^L - 1) & \text{if } i = y \;, \\
 		a_i^L a_y^L & \text{if } i \neq y \;.
 	\end{cases}
\end{equation}
The second gradient on the RHS of Eq~\eqref{eq:grad_cost3} is ill-defined, but can be approximated by making certain assumptions regarding the neuron's dynamics close to its firing threshold. Specifically, for a deterministic \ac{LIF} neuron it follows that the gradient of the neuron's membrane potential must be positive at its firing threshold when a spike is fired, such that $\partial u_i^L / \partial t(\tau_i) > 0$. Hence, following \cite{Bohte2002a}, we make a first order approximation of $u_i^L$ for a small region about $t=\tau_i$, giving rise to the relation $\delta \tau_i = -\delta u_i^L / \alpha$, where the local gradient is given by $\alpha = \partial u_i^L / \partial t(\tau_i)$. Taken together, the gradient of the neuron's first firing time is approximated by
\begin{align} \label{eq:grad_cost5}
	\frac{\tau_i}{\partial u_i^L} &\approx \frac{\partial \tau_i(u_i^L)}{\partial u_i^L(t)}\bigg\rvert_{u_i^L = \theta} \nonumber \\
		&\approx -\frac{1}{\partial u_i^L / \partial t}\bigg\rvert_{t = \tau_i} = -\frac{1}{\alpha} \;,
\end{align}
where for numerical stability reasons $\alpha$ is considered to be a positive, constant value. For the sake of brevity this constant is set to unity in the remainder of this analysis, and gives no qualitative change in the final result. Thus, Eqs~\eqref{eq:grad_cost2} to \eqref{eq:grad_cost5} are combined to give one of two possible output neuron error signals:
\begin{equation} \label{eq:grad_cost6}
\frac{\partial C(\mathbf{y}, \mathbf{a}^L)}{\partial u_i^L} = 
	\begin{cases}
		a_i^L - 1 & \text{if } i = y \;, \\
 		a_i^L & \text{if } i \neq y \;,
 	\end{cases}
\end{equation}
depending on whether the $i^\mathrm{th}$ neuron corresponds to the target label $y$. Using our earlier notation for the network's target activation vector $\mathbf{y} = (y_1, y_2, \dots, y_c)$ as used in Eq~\eqref{eq:cross-entropy}, the above can be written more compactly as
\begin{align} \label{eq:errorL}
\delta_i^L
	&:= \frac{\partial C(\mathbf{y}, \mathbf{a}^L)}{\partial u_i^L} \nonumber \\
	&:= a_i^L - y_i \;,
\end{align}
where we define $\delta_i^L$ to be the error signal due to the $i^\mathrm{th}$ neuron in the final layer.

\subsubsection{Output Weight Updates}

\begin{figure}[t]
\centering
\includegraphics{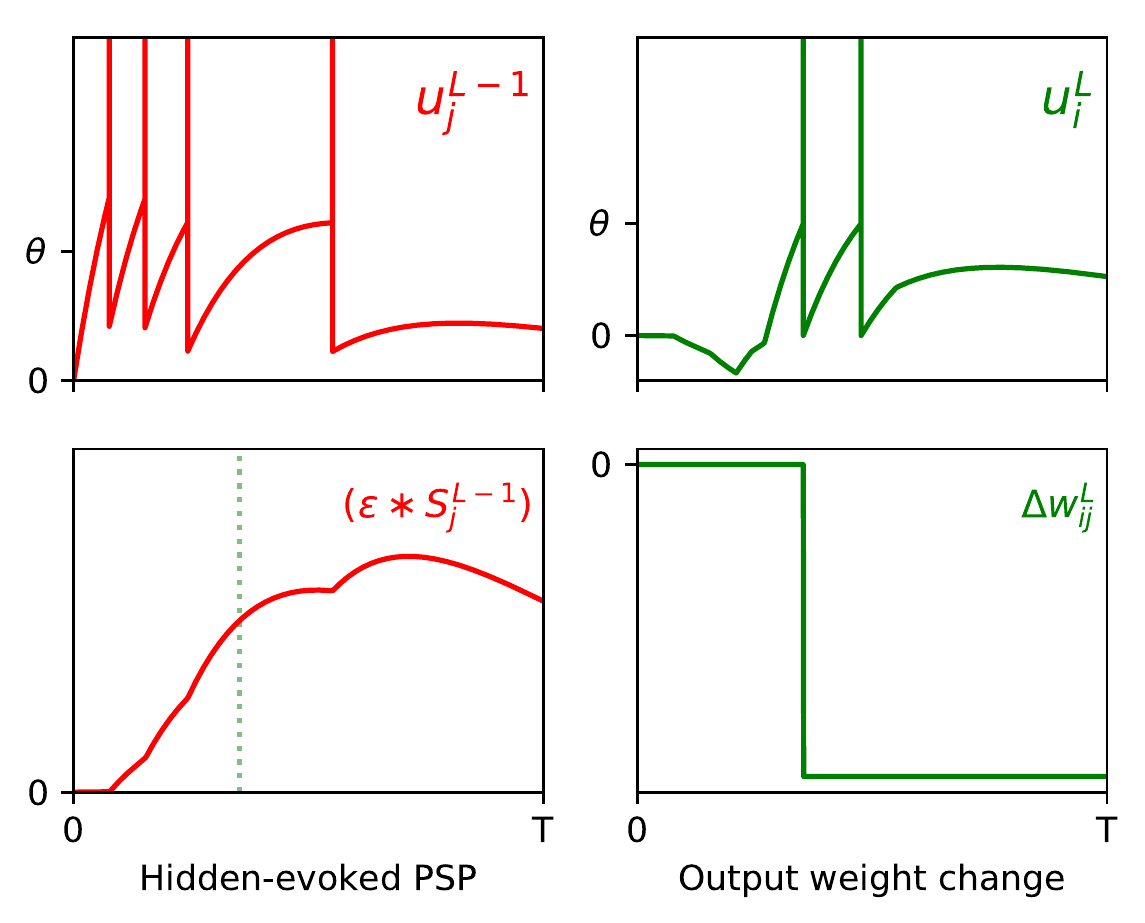}
\caption{
Example of the weight update process with respect to the output layer, $l=L$, of a multilayer \ac{SNN}, according to Eq~\eqref{eq:dw_L}. (Top row) The left panel shows the membrane potential of an excited presynaptic neuron $j$ in the second-to-last layer, $l=L-1$, over a small observation time $T$. The vertical lines indicate the neuron's firing times, and $\vartheta$ its firing threshold. The right panel shows the response of a postsynaptic neuron $i$ in the network's output layer, which is stimulated in part by neuron $j$. (Bottom row) The left panel shows the \ac{PSP} evoked due to neuron $j$. The first output spike fired by neuron $i$ is indicated by the green dotted line. Hence, as shown in the right panel, the magnitude of the weight change between neurons $j$ and $i$ is proportional to the value of the \ac{PSP} at the moment neuron $i$ fires its first spike. In this example the output neuron does not correspond to the class of the input pattern, therefore the direction of the weight change is negative to discourage early spiking.
}
\label{fig:output_dw}
\end{figure}

We apply gradient descent to Eq~\eqref{eq:cross-entropy-onehot} with respect to final layer weights, such that the weight between the $i^\mathrm{th}$ output neuron and $j^\mathrm{th}$ previous layer, hidden neuron is modified according to
\begin{equation} \label{eq:grad_wo}
\Delta w_{ij}^L = -\eta \frac{\partial C(\mathbf{y}, \mathbf{a}^L)}{\partial w_{ij}^L} \;,
\end{equation}
where $\eta > 0$ is the learning rate. The second term on the RHS is expanded using the chain rule to give
\begin{equation} \label{eq:grad_wo_2}
\frac{\partial C(\mathbf{y}, \mathbf{a}^L)}{\partial w_{ij}^L} = \frac{\partial C(\mathbf{y}, \mathbf{a}^L)}{\partial u_i^L} \frac{\partial u_i^L(t)}{\partial w_{ij}^L} \bigg\rvert_{t = \tau_i}  \;,
\end{equation}
where the gradient of the output neuron's membrane potential is evaluated at the time of its first spike. The first gradient term on the RHS of this equation corresponds to the neuron's error signal, as provided by Eq~\eqref{eq:errorL}, hence the above can be rewritten as
\begin{equation} \label{eq:grad_wo_3}
\frac{\partial C(\mathbf{y}, \mathbf{a}^L)}{\partial w_{ij}^L} = \delta_i^L \frac{\partial u_i^L(t)}{\partial w_{ij}^L} \bigg\rvert_{t = \tau_i}  \;.
\end{equation}
Using the definition of the neuron's membrane potential given by Eq~\eqref{eq:potential}, and neglecting the contribution due to refractory effects which is valid for sufficiently low output firing rates, the above becomes
\begin{equation} \label{eq:dw_L}
\Delta w_{ij}^L = -\eta \delta_i^L \left( \epsilon \ast S_j^{L-1} \right)(\tau_i) \;,
\end{equation}
where the constant $\alpha$, as introduced by Eq~\eqref{eq:grad_cost5}, is folded into $\eta$ for simplicity. An example of the above weight update rule taking place in a simulated \ac{SNN} is visualised in Fig~\ref{fig:output_dw}.

\paragraph{Integrated formula.} Integrating out the spike train in Eq~\eqref{eq:dw_L} gives
\begin{equation} \label{eq:dw_L_integrated}
\Delta w_{ij}^L = -\eta \delta_i^L \sum_{t_j^f \in \mathcal{F}_j^{L-1}} \epsilon (\tau_i - t_j^f) \;,
\end{equation}
where $\mathcal{F}_j^{L-1}$ is used to denote the list of spike times contributed by the $j^\mathrm{th}$ neuron in the previous layer, $L-1$.

\subsubsection{Hidden Weight Updates}

\begin{figure}[!tp]
\centering
\includegraphics[width=\textwidth]{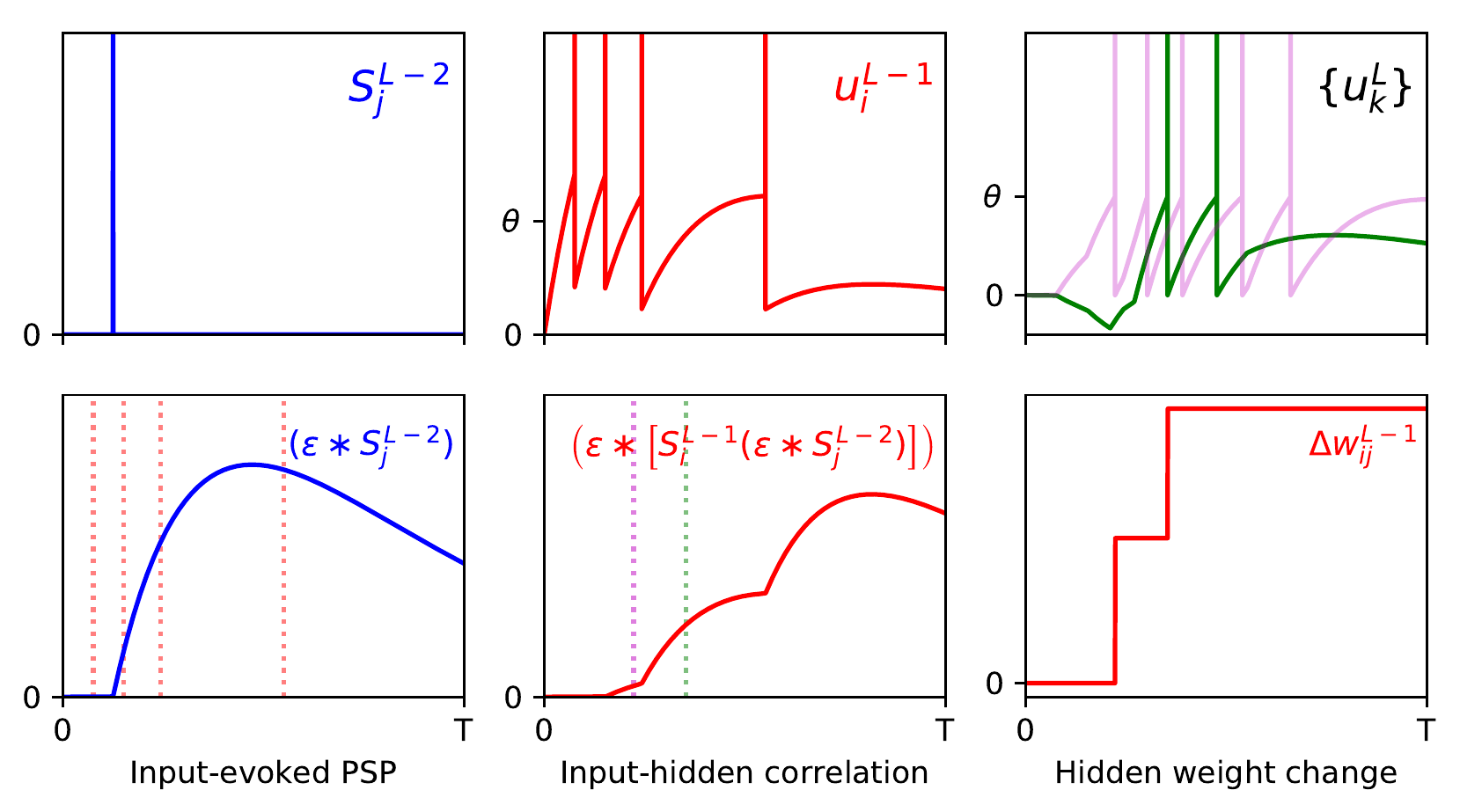}
\caption{
Example of the weight update process with respect to a hidden layer, $l=L-1$, of a multilayer \ac{SNN}, according to Eq~\eqref{eq:dw_L-1}. (Top row) The left panel shows the spike train of a presynaptic neuron $j$ in the first layer, $l=L-2$, observed over $T$. As shown in the middle panel, a postsynaptic neuron $i$ in layer $l=L-1$ receives this input spike, among others, and responds by firing a sequence of hidden spikes. The right panel shows the response of all output neurons in the network, each of which are stimulated in part by neuron $i$. (Bottom row) The left panel shows the \ac{PSP} evoked by input neuron $j$ at its postsynaptic target $i$. The red dotted lines indicate the firing times of $i$, which capture input-to-hidden spike correlations as shown in the middle panel. Since the first hidden spike occurs before the input spike, this spike makes no contribution to the input-hidden correlation. The middle panel also indicates the first firing times of all neurons in the output layer, marked by the magenta (first output to respond) and green dotted lines. The values of the input-hidden correlation trace at these two moments are used to inform the candidate weight change between input and hidden neurons $j$ and $i$, as shown in the right panel. In this example, the output neuron corresponding to the correct class label (magenta) is the first to fire a spike, as desired. The hidden neuron $i$ projects positive- and negative-valued synaptic weights to the magenta and green output neurons, respectively, hence this translates into a positive change in $w_{ij}^{L-1}$ at each moment.
}
\label{fig:hidden_dw}
\end{figure}

With respect to hidden layer weight updates, gradient descent is taken on Eq~\eqref{eq:cross-entropy-onehot} according to
\begin{equation} \label{eq:grad_wh}
\Delta w_{ij}^{L-1} = -\eta \frac{\partial C(\mathbf{y}, \mathbf{a}^L)}{\partial w_{ij}^{L-1}} \;,
\end{equation}
where the weight update between the $i^\mathrm{th}$ hidden neuron in layer $L-1$ and the $j^\mathrm{th}$ presynaptic neuron in layer $L-2$ is derived. Hence, using the chain rule, the gradient on the RHS is expanded as follows:
\begin{align} \label{eq:grad_wh_2}
\frac{\partial C(\mathbf{y}, \mathbf{a}^L)}{\partial w_{ij}^{L-1}}
	&= \sum_{k \in \Gamma^{i, L-1}} \frac{\partial C(\mathbf{y}, \mathbf{a}^L)}{\partial u_{k}^L} \frac{\partial u_{k}^L(t)}{\partial w_{ij}^{L-1}} \bigg\rvert_{t = \tau_{k}} \nonumber \\
	&= \sum_{k \in \Gamma^{i, L-1}} \delta_{k}^L \frac{\partial u_{k}^L(t)}{\partial w_{ij}^{L-1}} \bigg\rvert_{t = \tau_{k}} \;,
\end{align}
where $\Gamma^{i, L-1}$ denotes the immediate set of neural successors of neuron $i$ in layer $L-1$, or the set of output layer neurons, and having used the identity of the output error signal given by Eq~\eqref{eq:errorL}. Using Eq~\eqref{eq:potential}, the gradient of the $k^\mathrm{th}$ membrane potential becomes
\begin{align} \label{eq:grad_wh_3}
\frac{\partial u_{k}^L(t)}{\partial w_{ij}^{L-1}}\bigg\rvert_{t = \tau_{k}}
&= w_{ki}^L \frac{\partial}{\partial w_{ij}^{L-1}} \left( \epsilon \ast S_i^{L-1} \right)(\tau_k) \nonumber \\
&= w_{ki}^L \left( \epsilon \ast \frac{\partial S_i^{L-1}}{\partial w_{ij}^{L-1}} \right)(\tau_k) \;,
\end{align}
where we neglect the contribution from the refractory term. Evaluating the gradient of a spike train poses a challenge given its discontinuous nature when generated by \ac{LIF} neurons. One approach to resolving this might instead just consider the first spike contributed by hidden layer neurons, as used for SpikeProp \citep{Bohte2002a}, although this loses information about neural firing frequency and typically requires the addition of multiple subconnections with the next layer to support sufficient downstream activation. There have been extensions of SpikeProp to allow for multiple spikes in the hidden layers \citep{Booij2005,Ghosh2009}, although these methods rely on small learning rates and constrained weight gradients to allow for convergence. To circumvent this issue, we treat hidden layer neurons as being probabilistic in order to provide smoother gradients \citep{Gardner2015}. Specifically, we introduce stochastic spike generation for hidden neurons using the Escape Noise model \citep{Gerstner2014}. By this mechanism, hidden neuron firing events are distributed according to an inhomogeneous Poisson process with a time-dependent rate parameter that is a function of the neuron's membrane potential: $\rho_i^l(t) = g(u_i^l(t))$. This can be interpreted as the neuron's instantaneous firing rate, or firing density, where the probability of the neuron firing a spike over an infinitesimal time window $\delta t$ is given by $\rho_i^l(t) \delta t$. Here we take an exponential dependence of the firing density on the distance between the neuron's membrane potential and threshold \citep{Gerstner2014}:
\begin{equation} \label{eq:escape_rate}
g(u_i^l(t)) = \rho_0 \exp \left( \frac{u_i^l(t) - \vartheta}{\Delta u} \right) \;,
\end{equation}
where $\rho_0 = \SI{0.01}{ms^{-1}}$ is the instantaneous rate at threshold, and $\Delta u = \SI{1}{mV}$ controls the variability of generated spikes. Hence, following our previous method in \cite{Gardner2015}, we can substitute the gradient of the spike train in Eq~\eqref{eq:grad_wh_3} with the gradient of its expected value, conditioned on spike trains in the previous layer of the network, such that
\begin{equation} \label{eq:grad_wh_4}
\frac{\partial S_i^{L-1}(t)}{\partial w_{ij}^{L-1}} \rightarrow \frac{\partial \left\langle S_i^{L-1}(t) \right\rangle_{S_i^{L-1}|\{S_j^{L-2}\}}}{\partial w_{ij}^{L-1}} \;.
\end{equation}
If we also condition the expected spike train on the neuron's most recently fired spike, $\hat{t}_i < t$, then we can express Eq~\eqref{eq:grad_wh_4} as the gradient of the instantaneous value of the spike train, distributed according to its firing density \citep{Fremaux2013}:
\begin{align} \label{eq:grad_wh_5}
\frac{\partial \left\langle S_i^{L-1}(t) \right\rangle_{S_i^{L-1}|\{S_j^{L-2}\}, \hat{t}_i}}{\partial w_{ij}^{L-1}}
&= \frac{\partial}{\partial w_{ij}^{L-1}} \sum_{q \in \{0,\delta(t)\}} q(t) \rho_i^{L-1}(t|\{S_j^{L-2}\}, \hat{t}_i) \nonumber \\
&= \delta_D(t-\hat{t}) \frac{\partial \rho_i^{L-1}(t|\{S_j^{L-2}\}, \hat{t}_i)}{\partial w_{ij}^{L-1}} \;,
\end{align}
where $\delta_D(t-\hat{t})$ is the Dirac-delta function centred on some most recent spike time $\hat{t}$. Using Eqs.~\eqref{eq:potential} and \eqref{eq:escape_rate}, and denoting `$|\{S_j^{L-2}\}, \hat{t}_i$' as `$|L-2, i$' for brevity, we obtain
\begin{align} \label{eq:grad_wh_6}
\frac{\partial \left\langle S_i^{L-1}(t) \right\rangle_{S_i^{L-1}|L-2, i}}{\partial w_{ij}^{L-1}}
&= \frac{1}{\Delta u} \delta_D(t-\hat{t}) \rho_i^{L-1}(t|L-2, i) \left( \epsilon \ast S_j^{L-2} \right)(t) \nonumber \\
&= \frac{1}{\Delta u} \left\langle S_i^{L-1}(t) \left( \epsilon \ast S_j^{L-2} \right)(t) \right\rangle_{S_i^{L-1}|L-2, i} \;.
\end{align}
We can estimate the expected value of the spike train's gradient through samples generated by the network during simulations, hence the above can be approximated as
\begin{equation} \label{eq:grad_wh_7}
\frac{\partial \left\langle S_i^{L-1}(t) \right\rangle_{S_i^{L-1}|L-2, i}}{\partial w_{ij}^{L-1}} \approx \frac{1}{\Delta u} S_i^{L-1}(t) \left( \epsilon \ast S_j^{L-2} \right)(t) \;.
\end{equation}
Combining Eqs.~\eqref{eq:grad_wh_3}, \eqref{eq:grad_wh_4} and \eqref{eq:grad_wh_7} provides an estimate for the gradient of the $k^\mathrm{th}$ output neuron's membrane potential, evaluated at the time of its first fired spike:
\begin{equation} \label{eq:grad_wh_8}
\frac{\partial u_{k}^L(t)}{\partial w_{ij}^{L-1}}\bigg\rvert_{t = \tau_{k}} = \frac{1}{\Delta u} w_{ki}^L \left( \epsilon \ast \left[ S_i^{L-1} \left( \epsilon \ast S_j^{L-2} \right) \right] \right)(\tau_k) \;.
\end{equation}
Hence, combining the above with Eqs.~\eqref{eq:grad_wh} and \eqref{eq:grad_wh_2} gives the second-last layer weight update rule:
\begin{equation} \label{eq:dw_L-1}
\Delta w_{ij}^{L-1} = -\frac{\eta}{\Delta u} \sum_{k \in \Gamma^{i, L-1}} \delta_{k}^L w_{ki}^L \left( \epsilon \ast \left[ S_i^{L-1} \left( \epsilon \ast S_j^{L-2} \right) \right] \right)(\tau_k) \;.
\end{equation}
An example of a weight update taking place between the input and hidden layers of an \ac{SNN} is shown in Fig~\ref{fig:hidden_dw}.

\paragraph{Integrated formula.} Integrating out the spike trains in Eq~\eqref{eq:dw_L-1} gives
\begin{equation} \label{eq:dw_L-1_integrated}
\Delta w_{ij}^{L-1} = -\frac{\eta}{\Delta u} \sum_{k \in \Gamma^{i, L-1}} \delta_k^L w_{ki}^L \sum_{t_i^f \in \mathcal{F}_i^{L-1}} \epsilon (\tau_k - t_i^f) \sum_{t_j^g \in \mathcal{F}_j^{L-2}} \epsilon (t_i^f - t_j^g) \;,
\end{equation}
where $\mathcal{F}_i^{L-1}$ and $\mathcal{F}_j^{L-2}$ are the list of spike times from neurons $i$ and $j$ in layers $L-1$ and $L-2$, respectively. The weight update formulae described by Eqs.~\eqref{eq:dw_L_integrated} and \eqref{eq:dw_L-1_integrated} determine the supervised learning process of our first-to-spike neural classifier rule, as applied to multilayer \acp{SNN} containing a single hidden layer of stochastic neurons. The above procedure is not restricted to \acp{SNN} containing one hidden layer, however: as demonstrated in the Appendix, it is also possible to extend this approach to networks containing more than one hidden layer.

\subsection{Temporal Encoding} \label{subsec:temporal_encoding}

For demonstrative purposes, this article studies the performance of the proposed multilayer learning rule as applied to a selection of benchmark classification datasets. To this end, it was necessary to first convert input features into spike-based representations: to be conveyed by the input layer of an \ac{SNN} for downstream processing. Therefore, we made use of three distinct encoding strategies to achieve this, including: latency-based, receptive fields and scanline encoding. An overview of each strategy is described as follows.

\subsubsection{Latency Encoding}
A straightforward means to forming a temporal representation of an input feature is to signal its intensity based on the latency of a spike. Specifically, if we consider an encoding \ac{LIF} neuron that is injected with a fixed input current, then the time taken for it to respond with a spike can be determined as a function of the current's intensity: by interpreting the feature's value as a current, it is therefore possible for it to be mapped to a unique firing time. For an encoding \ac{LIF} neuron $i$ with a fixed firing threshold that only receives a constant current $I_i$, its first response time is given by \citep{Gerstner2002}:
\begin{equation} \label{eq:response_time}
t_i^1 = 
	\begin{cases}
		\tau_m \log \left( \frac{R I_i}{R I_i - \vartheta} \right) & \text{if } R I_i > \vartheta \;, \\
 		\infty & \text{otherwise} \;,
 	\end{cases}
\end{equation}
where we use the same parameter selections for $\tau_m$ and $\vartheta$ as used in section~\ref{subsec:nrn_model}, and the resistance is set to $R = \SI{4}{M \Omega}$. In terms of relating feature values to current intensities, we take one of two different approaches. In the first approach we arbitrarily associate each feature value with a unique intensity value, which is ideally suited to the case where features are limited to a small number of discrete values. In the second approach, and in the case where features take real values, we devise a more direct association; specifically, each value $x_i$ belonging to a feature vector $\mathbf{x}$ is normalised to fall within the unit range before being scaled by a factor $I_\mathrm{max}$, providing the current intensity $I_i$. The specific choice of $I_\mathrm{max}$ used depends on the studied dataset. Regardless of the approach we take, and in order to maintain a tight distribution of early spike arrivals, we disregard spikes with timings greater than \SI{9}{ms} by setting them to infinity.

\subsubsection{Receptive Fields}
An alternative, population-based approach to encoding real-valued variables relies on the concept of receptive fields. This strategy was first described by \cite{Bohte2002a} in the context of \acp{SNN}, and involves setting up a population of neurons with overlapping, graded response curves which are individually sensitive to a certain subset of values an encoded feature can take. Typically, a Gaussian-shaped response curve (or receptive field) is assumed, where an early (late) spike fired by a neuron corresponds to strong (weak) overlap with its encoded feature. For the datasets that are encoded in this way, we assign $q$ neurons with Gaussian receptive fields to each feature. For the $i^\mathrm{th}$ feature, with values existing in the range $[x_i^\mathrm{min}, \dots, x_i^\mathrm{max}]$, its encoding neural fields are centred according to $x_i^\mathrm{min} + (2 j - 3) / 2 \cdot (x_i^\mathrm{max} - x_i^\mathrm{min}) / (q - 2)$, for encoder indices $1 \leq j \leq q$, and using the width parameter $\sigma_i = 2 / 3 \cdot (x_i^\mathrm{max} - x_i^\mathrm{min}) / (q - 2)$ \citep{Bohte2002a}. Hence, a data sample consisting of $n_f$ features results in a matrix of first-layer neural activations: $\mathbf{a}^1 \in \mathbb{R}^{n_f \times q}$, with values in the range $(0, 1]$. As in \cite{Bohte2002a}, these activations are then mapped to a matrix of single spike time encodings according to $\mathbf{t}^1 = 10 \cdot (1 - \mathbf{a}^1)$, where values greater than \num{9} are discarded since they are deemed insufficiently activated. For the datasets transformed by receptive fields in this article, we used a different number of encoding neurons to give the best performance.

\subsubsection{Scanline Encoding}
A promising strategy for transforming visual data into spike patterns is to apply `scanline encoding', a technique that has been described in \cite{Lin2018}. Scanline encoding is a method inspired by human saccadic eye movements, and works to detect object edges within images when scanned across multiple line projections; when an increase in pixel intensity is detected along one of these scanlines, an associated, encoding spike is generated. The efficiency of this method derives from its subsampling of image pixels using a limited number of scanlines, as well as its invariance to small, local image distortions; in this way, it is possible to perform dimensionality reduction on images with spatial redundancy. There are several ways in which scanline encoding can be implemented, and the specific approach taken by \cite{Lin2018} represents just one possible choice. In general, the first step involves setting up a number of scanlines with certain orientations that are fixed with respect to all training and test images; ideally, the directions of these scanlines should be selected to capture the most informative pixels, as determined through preliminary parameter sweeps. The next step is to then preprocess the images by reading the values of pixels along these directions. In \cite{Lin2018} these scanlines are additionally split into several segments, where each segment maps to a channel that injects spikes into an \ac{SNN}: if a contrast shift is detected anywhere along one of these segments, then a repeating spike train is generated for the corresponding channel. Although using segmented-scanlines provided the authors with a test accuracy of \SI{96.4}{\%} on MNIST, a large number of neurons was required to encode the images in this way and the significance of individual spike timings was disregarded.

\begin{figure}[t]
\centering
\includegraphics{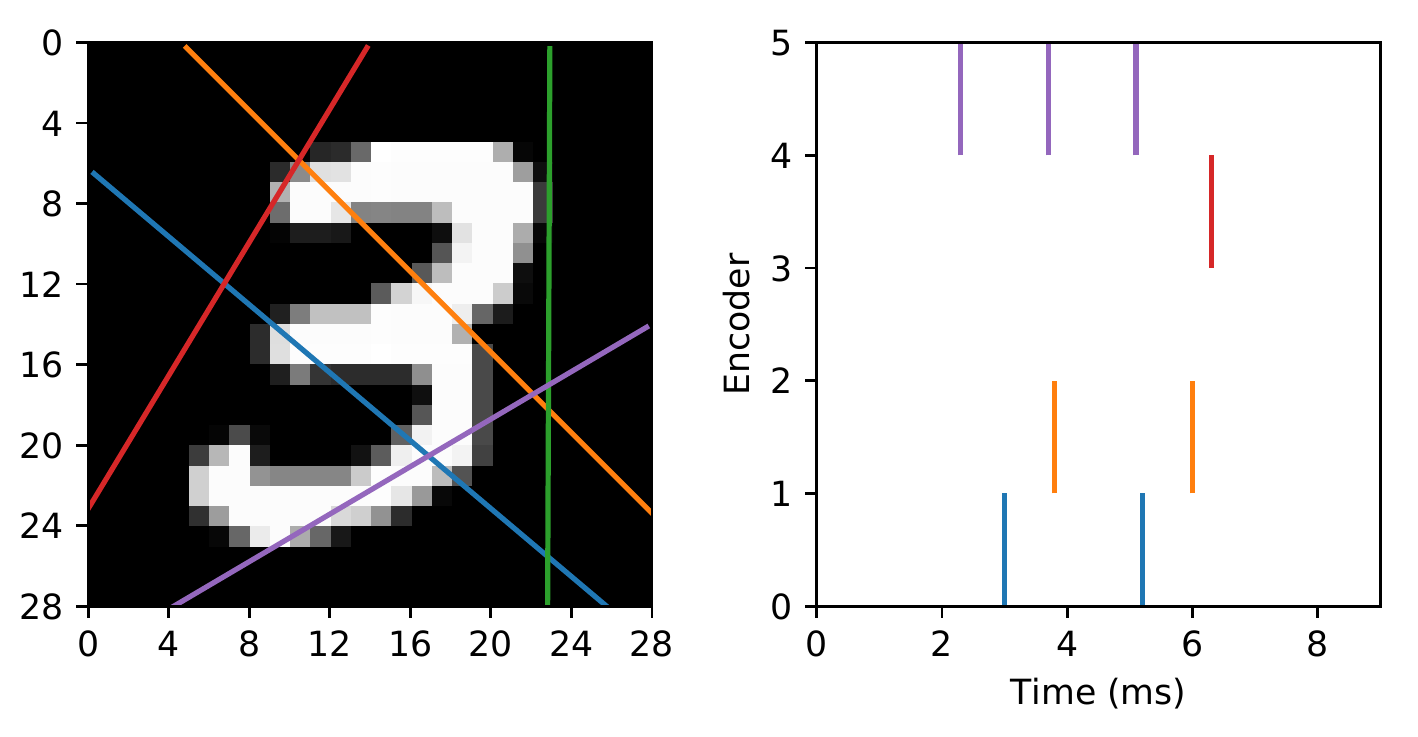}
\caption{
Example of an MNIST digit being transformed into five spike trains using `scanline encoders'. In this case, the orientations of the scanlines were randomly predetermined, and pixel intensities lying underneath each line were read-in working from the bottom upwards over \SI{9}{ms}.
}
\label{fig:scanline_encoding}
\end{figure}

In our approach we wish to fully utilise the timings of individual spikes to maintain sparse image  representations, and also to avoid artificially segmenting scanlines in the first instance. To this end, we modify the spike generation process by instead interpreting the read-in pixel values underneath each line as a series of sequentially-occurring current stimulus values. Hence, if we assume these values are injected over some duration to an encoding \ac{LIF} neuron, then we arrive at a sequence of precisely-timed spikes representing each scanline. In terms of parameter selection, the encoding \ac{LIF} neurons are designed to be relatively fast responders: with their membrane time constants set to \SI{3}{ms}. The resistance of each neuron is set to $R = \SI{10}{M \ohm}$, with a firing threshold of just one millivolt to elicit a rapid response. Immediately after firing a spike an encoding neuron's membrane potential is reset to \SI{0}{mV}, and the neuron is forced to remain quiescent for \SI{1}{ms}. With respect to the scanlines, we first decide on a number $n_s$ according to the experimental design. Each scanline is then setup as follows. First, the orientation of each line is selected according to a uniform distribution over the range $[0, \pi)$. Each line is then set to intercept through a position that is normally-distributed about the image centre, where the scale parameter of this distribution is a quarter of the image width. These scanlines remain fixed across all training and test images. Hence, when an image is encoded, the pixels lying underneath a scanline are injected as current stimulus values into a corresponding \ac{LIF} encoder, after first normalising pixels to exist in the unit range. Pixels are always scanned-in from the bottom of an image upwards, over a duration of \SI{9}{ms}. An example of this encoding strategy is illustrated in Fig~\ref{fig:scanline_encoding}.

\subsection{Network Structure} \label{subsec:network_structure}

In all of the experiments we considered fully-connected, feedforward \acp{SNN} containing a single hidden layer of spiking neurons, such that $L=3$. Data samples presented to a network were encoded by the collective firing activity of input layer neurons, according to one of the temporal encoding strategies described above; hidden layer neurons were free to perform computations on these input patterns, and learn features useful for downstream processing. Neurons in the last, or output, layer of a network were tasked with forming class predictions on these data samples according to a first-to-spike mechanism, where the predicted class label was determined according to which one of $N_{l=3} = c$ output neurons was the first to respond with an output spike. The number of neurons implemented in the input and hidden layers, $N_1$ and $N_2$, respectively, depended on the type of input data and the run experiment, although we generally aimed to design a minimalistic setup for efficiency reasons. As described in section~\ref{subsec:learning_rule}, stochastic and deterministic \ac{SRM} neurons were simulated in the hidden and output layers, respectively, and, unless otherwise stated, all neurons within a layer shared the same cellular parameters. For all experiments, the internal simulation time step was set to $\delta t = \SI{0.1}{ms}$.

In terms of network connections, we initialised hidden and output weights to drive the initial firing rates of neurons to approximately one spike in response to presented data samples. Initial weight values were drawn from a uniform distribution, as detailed in the description of each experiment. Unless otherwise stated, each pre- and postsynaptic neuron pair had a single connection with no conduction delay in spike propagation.

\subsection{Synaptic Plasticity} \label{subsec:synaptic_plasticity}

Synaptic weights projecting onto the hidden and output layers of multilayer \acp{SNN} were modified during training via a combination of synaptic plasticity rules, while subject to certain constraints. This process is described in detail as follows.

\subsubsection{Learning Procedure}

During training, data samples were presented to a network iteratively as mini-batches, where computed weight gradients were accumulated over individual samples before actually being applied at the end of each mini-batch; this procedure was selected in order to reduce the variance of weight changes in a network to provide smoother convergence, as well as to obtain more reliable estimates of network activity as needed for regularisation. The order in which data samples were assigned as mini-batches was random, and, unless otherwise specified, the number of samples in each batch was 150. Furthermore, weight gradients were only computed after observing the entire response of a network when stimulated by a data sample, which in most cases was completed after duration $T = \SI{40}{ms}$ given input spikes arriving within \SI{10}{ms} of pattern onset. Hence, if the network was presented with a data sample described by an input vector $\mathbf{x}$ and a one-hot encoded class label $\mathbf{y}$, then, after applying a suitable temporal encoding strategy, a synaptic weight gradient in the $l^\mathrm{th}$ layer was determined as
\begin{equation} \label{eq:weight_update}
\Delta w_{ij}^l = -\eta \left( \frac{\partial C(\mathbf{y}, \mathbf{a}^L)}{\partial w_{ij}^l} + \lambda(S_i^l) - \gamma(S_i^l) \right) \;,
\end{equation}
where the first term in brackets on the RHS is the gradient of the cost function, which is evaluated following the steps of Eq~\eqref{eq:grad_wo} or \eqref{eq:grad_wh} for the output and hidden layers, respectively. The second term, $\lambda(S_i^l)$, is a regularisation function which depends on the postsynaptic neuron's firing activity, and the final term, $\gamma(S_i^l)$, is a synaptic scaling function. These two functions are defined as follows.

\subsubsection{Regularisation Term}

As a means to encourage network generalisation we enforced an L2 penalty term with respect to hidden and output layer weight gradients during training. Additionally, we also included a factor penalising high neuronal firing rates: a strategy that has been demonstrated in \cite{Zenke2018} to provide increased network stability as well as boosted performance. The regularisation term is defined by
\begin{equation} \label{eq:regularisation}
\lambda(S_i^l) = \lambda_0 w_{ij}^l \zeta(S_i^l) \;,
\end{equation}
where $\lambda_0$ is a scaling factor that is optimised for each experiment and the function $\zeta(S_i^l) = \left[ \int_{t=0}^{T} S_i^l(t) \mathrm{d}t \right]^2$ is an activity-dependent penalty term that depends on the number of spikes fired by a neuron $i$ in layer $l$. Since data samples were iteratively presented to the network, and the observation period $T$ was set sufficiently large, integrating over a spike train accurately reflected a neuron's firing rate. Through preliminary simulations we found that selecting an exponent greater than one with respect to the dependence of $\zeta$ on a neuron's firing rate gave the best results, consistent with that found in \cite{Zenke2018}.

\subsubsection{Synaptic Scaling Term}

It was necessary to include a synaptic scaling term as part of the weight gradient computation in order to sustain at least a minimum of activity in the network during training. This is because the weight update formulae described by Eqs~\eqref{eq:dw_L} and \eqref{eq:dw_L-1} both depend on presynaptic spikes in order to compute output and hidden weight gradients, which would result in non-convergent learning if no spikes could be  acted upon. Adapting the synaptic scaling rule described in our previous work \citep{Gardner2015}, as well as taking inspiration from the scaling procedure used in \cite{Mostafa2017}, we define the synaptic scaling term as follows:
\begin{equation} \label{eq:syn_scaling}
\gamma(S_i^l) = 
	\begin{cases}
		\gamma_0 |w_{ij}^l| & \text{if } \int_{t=0}^{T} S_i^l(t) \mathrm{d}t = 0 \;, \\
 		0 & \text{otherwise}\;,
 	\end{cases}
\end{equation}
where $\gamma_0 = 0.1$ is a scaling parameter. From a biological perspective, synaptic scaling can be interpreted as a homoeostatic learning factor that assists with maintaining desired activity levels \citep{vanRossum2000}.

\subsubsection{Learning Schedule}

The learning procedure used to compute weight gradients, defined by Eq~\eqref{eq:weight_update}, was accumulated over all data samples assigned to a mini-batch before weights were actually updated in a trained network. However, rather than directly using these computed gradients, we took the additional step of applying synapse-specific, adaptive learning rates to modulate the magnitude of the weight updates. As found in \cite{Zenke2018}, and through preliminary simulations, we found that a technique referred to as RMSProp \citep{Hinton2012} was more effective in providing convergent performance than applying a global, non-adaptive learning rate, and proved less sensitive to the experimental design. Specifically, an accumulated weight gradient $\Delta w_{ij}^l$, as determined using Eq~\eqref{eq:weight_update}, was used to inform a weight update via RMSProp according to
\begin{align} \label{eq:RMSProp}
w_{ij}^l
&\leftarrow w_{ij}^l + \frac{\eta_0}{\sqrt{m_{ij}^l + \varepsilon}} \Delta w_{ij}^l \nonumber \\
m_{ij}^l
&\leftarrow \beta m_{ij}^l + (1 - \beta) \left( \Delta w_{ij}^l \right)^2 \;,
\end{align}
where $\eta_0 > 0$ is a constant coefficient that was specific to each experiment, $m_{ij}^l$ is an auxiliary variable that keeps track of the recent weight gradient magnitudes, $\varepsilon = \num{1E-8}$ is a small offset that was included for numerical stability and $\beta = 0.9$ is a decay factor. The initial value of $m_{ij}^l$ was taken to be zero. Additionally, weights were constrained to a range, $w_{ij}^l \in [w_\mathrm{min}, w_\mathrm{max}]$, in order to prevent overlearning during training. The weight limits were specific to each of the studied experiments.

\section{Results}

\subsection{Solving the XOR Task} \label{subsec:xor_task}

As a first step in assessing the performance of the first-to-spike multilayer classifier, we tested its ability to classify data samples that were linearly non-separable. A classic benchmark for this is the \ac{XOR} classification task, a non-trivial problem for which a hidden layer of spiking neurons is indicated to be necessary for its solution \citep{Gruning2012,Gardner2015}.

\begin{figure}[!tp]
\centering
\includegraphics[width=\textwidth]{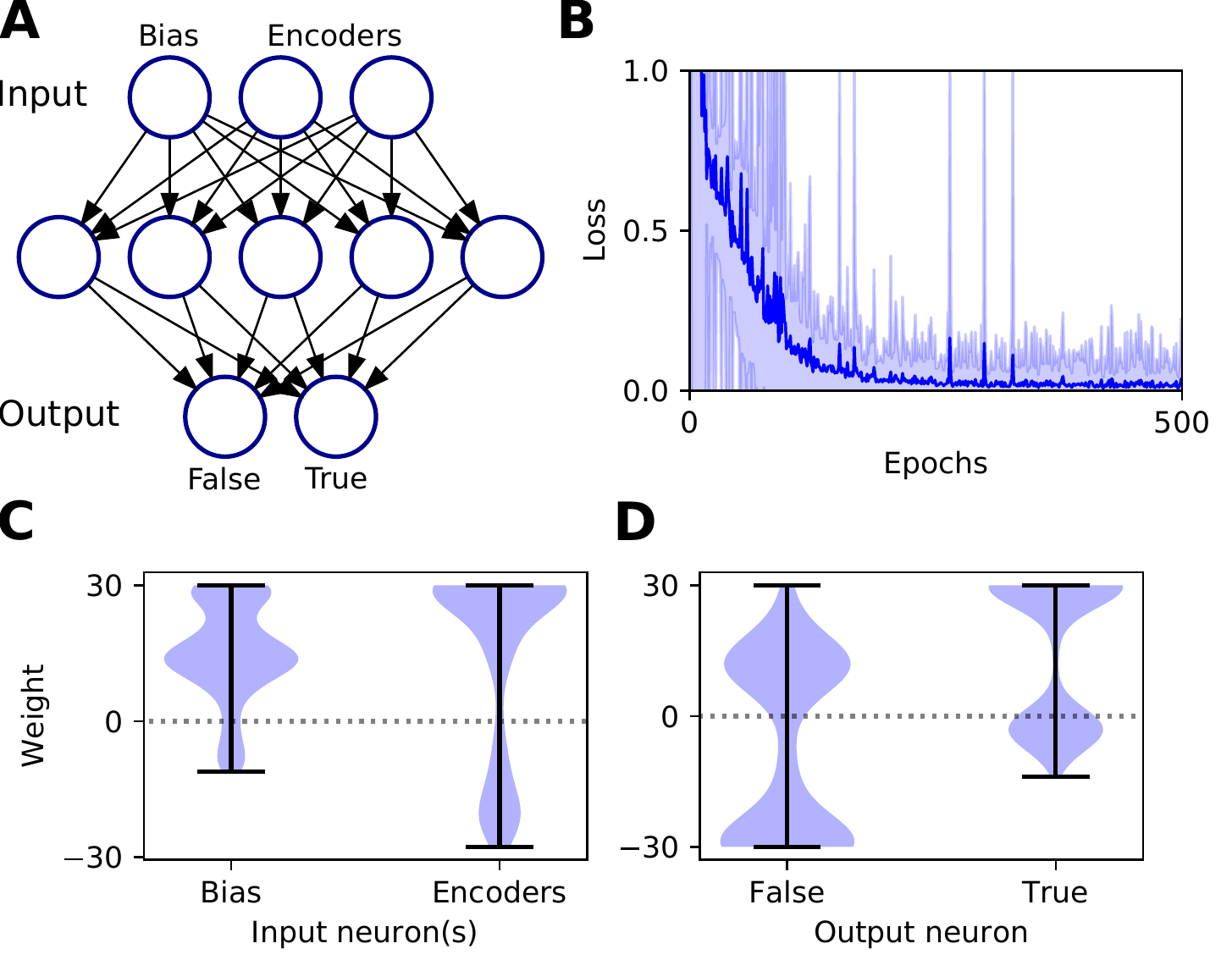}
\caption{
Solving the \ac{XOR} computation in a multilayer \ac{SNN} using first-to-spike decoding. (A) Network setup: consisting of three, five and two neurons in the input, hidden and output layers, respectively. Input patterns were encoded by the latencies of spikes in the input layer, and their class labels were predicted based on which one of the two output neurons was the first to respond with a spike. (B) Network loss as a function of the number of training epochs, averaged over 100 independent runs. The shaded region shows the standard deviation. (C) Post-training hidden layer weight distributions. This panel breaks down the overall distribution of hidden weights into two components: one with respect to connections projecting from the input bias neuron, and the other due to input encoder neurons. The shaded width corresponds to the probability density of weights. (D) Post-training output layer weight distributions. This panel shows the distribution of output weights with respect to connections projecting from the hidden layer onto False- or True-signalling output neurons.
}
\label{fig:xor_results}
\end{figure}

An \ac{XOR} computation takes as its input two binary-valued input variables, and maps them to a single binary target output in the following way: $\{0, 0\} \rightarrow 0$, $\{0, 1\} \rightarrow 1$, $\{1, 0\} \rightarrow 1$ and $\{1, 1\} \rightarrow 0$, where 1 and 0 correspond to Boolean True and False, respectively. To make this scheme compatible with \ac{SNN} processing, we first transformed the input values into spike-based representations using an appropriate temporal encoding strategy. In this case, each binary value was encoded by the latency of an input spike, such that values of 1 and 0 corresponded to early and late spike timings, respectively. For simplicity, we selected the associated current intensities, as defined by Eq~\eqref{eq:response_time}, such that an input value of 1 resulted in a spike latency of \SI{0}{ms}, and an input value of 0 resulted in a spike latency of \SI{6}{ms}. Furthermore, in order to make the task non-trivial to solve, and to allow the network to discriminate between input patterns presenting both True or False values, we also included an input bias neuron that always fired a spike at \SI{0}{ms} to signal pattern onset \citep{Bohte2002a}. Hence, as illustrated in Fig~\ref{fig:xor_results}A, we setup an \ac{SNN} which contained three input neurons (one bias and two encoders), five hidden neurons and two output neurons to signal the two possible class labels (True / False). At the start of each experiment run, hidden and output weights were initialised by drawing their values from uniform distributions over the ranges: $[0, 16)$ and $[0, 6.4)$, respectively. The softmax scale parameter defining output activations was set to $\nu = 2$. In terms of network training, results were gathered from runs lasting 500 epochs, where each epoch corresponded to the presentation of all four input patterns. Regularisation was not required on this task, so we set $\lambda_0 = 0$. The RMSProp coefficient was set to $\eta_0 = 0.5$, and throughout training hidden and output weights were constrained between $[-30, 30]$.

\begin{figure}[!tp]
\centering
\includegraphics[width=\textwidth]{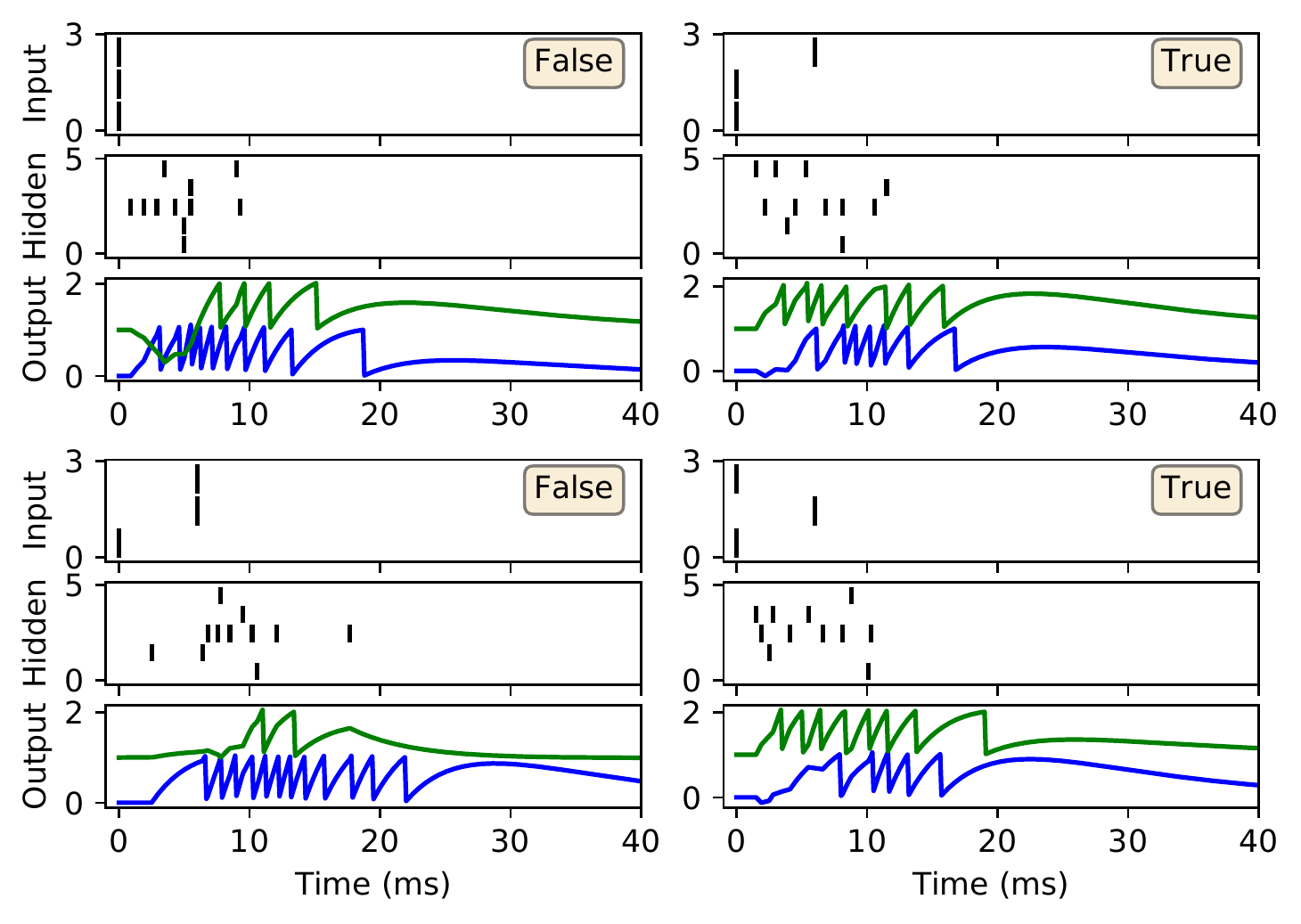}
\caption{
Spike rasters depicting network activity on the \ac{XOR} task, with the same setup as in Fig~\ref{fig:xor_results}, after 200 epochs of training. Each of these four main panels depicts network activity in response to one of the four possible \ac{XOR} input patterns, where the associated class label is indicated as either `True' or `False'. Each main panel conists of three subplots: the top and middle subplots correspond to input and hidden spike patterns, respectively, where vertical lines indicate spikes. The bottom subplot displays the output layer voltage traces, where firing times are indicated by sharp discontinuities in these traces. Of the two output neurons, the first (blue trace) signals the class label `False', and the second (green trace) signals `True'. In this example, all of the input patterns were correctly identified.
}
\label{fig:xor_rasters}
\end{figure}

As shown in Fig~\ref{fig:xor_results}B, the network was successful in learning the \ac{XOR} task: reaching a final training loss of \num{0.02 \pm 0.01} as obtained from 100 independent runs (error reported as \ac{SEM}). This reflected a final classification accuracy of \num{99.8 \pm 0.2}{\%}, which didn't reach precisely \num{100}{\%} due to the stochastic nature of hidden layer spike generation. In terms of the final weight distributions of the network (Fig~\ref{fig:xor_results}C, D), systematic trends were observed for certain connections in the hidden and output layers. With respect to the hidden layer, incoming connections received from the encoder neurons were widely distributed, with just over \num{70}{\%} being excitatory. By comparison, the bias neuron tended to project positively-skewed weights, with almost \num{90}{\%} being excitatory; the relatively large fraction of excitatory connections indicated its role in sustaining hidden layer activity, irrespective of the input pattern statistics. With respect to the final distribution of output layer weights, the False- and True-signalling neurons differed from each other by assuming a greater proportion of weight values saturating towards their lower and upper limits, respectively. The result of this distribution was to suppress the erroneous output neuron until the desired one received sufficient input activation to fire; this process can be inferred from the spike rasters depicted in Fig~\ref{fig:xor_rasters}, showing the hidden and output layer responses of a post-trained network when presented with each of the four input patterns. For example, when presented with patterns labelled as `False' the output neuron signalling `True' responded with a small number of delayed spikes, whereas the correct neuron promptly responded with multiple, rapid spikes.

\subsection{Classifying the Iris and Wisconsin Datasets}

A key determinant of a classifier system's performance is its ability to generalise from training data to previously unseen test samples. In order to assess the generalisation ability of the proposed learning rule we applied it to classifying two benchmark datasets: Iris and Wisconsin.

The Iris dataset \citep{Fisher1936} presents a multi-class classification problem, containing 150 data samples evenly split between three different classes: two of which are not linearly separable from each other. Each sample is described by four real-valued features measuring some of the attributes of three types of Iris flower. The Wisconsin breast cancer dataset \citep{Wolberg1990} is a binary classification problem, containing 699 samples split between two different classes. Each sample consists of nine discrete-valued features measuring the likelihood of malignancy based on their intensity, and is labelled as either benign or malignant. We note that of these 699 original samples 16 contained missing values, which we discarded to provide a revised total of 683. In terms of our strategy for transforming these two datasets into spike-based representations, we followed the approach of \cite{Bohte2002a}; specifically, Gaussian receptive fields were applied as a means to converting the input features into first-layer spike latencies, resulting in spike-timings distributed between 0 and \SI{9}{ms} (see section~\ref{subsec:temporal_encoding}). For Iris, consisting of $n_f = 4$ features, we assigned a population of $q = 12$ input neurons to encode each feature, resulting in a total input layer size $N_1 = 48$. For Wisconsin, with $n_f = 9$ features, a population of $q = 7$ neurons per feature was assigned, resulting in $N_1 = 63$ input neurons. As usual, one output neuron was assigned to each input class, and for both Iris and Wisconsin we implemented a hidden layer size of 20. Hence, the \ac{SNN} structures were described by $48 \times 20 \times 3$ and $63 \times 20 \times 2$ for Iris and Wisconsin, respectively. At the start of each experimental run for Iris and Wisconsin, output weights were initialised with values drawn from a uniform distribution over $[0, 2)$. Hidden weights were initialised according to uniform distributions over $[0, 4)$ and $[0, 2.2)$ for Iris and Wisconsin, respectively. For both datasets the softmax scale parameter of Eq~\eqref{eq:softmax} was set to $\nu = 2$. With respect to network training, stratified three-fold cross-validation was used to obtain more accurate estimates for the network performance. Data samples were presented to the network as mini-batches, and one epoch of training corresponded to a complete sweep over all unique training samples presented in this way. The regularisation parameter was set to $\lambda_0 = \num{E-3}$ and the RMSProp coefficient: $\eta_0 = 0.1$. In all cases, hidden and output weights were constrained to values in the range $[-15, 15]$.

\begin{figure}[!tp]
\centering
\includegraphics{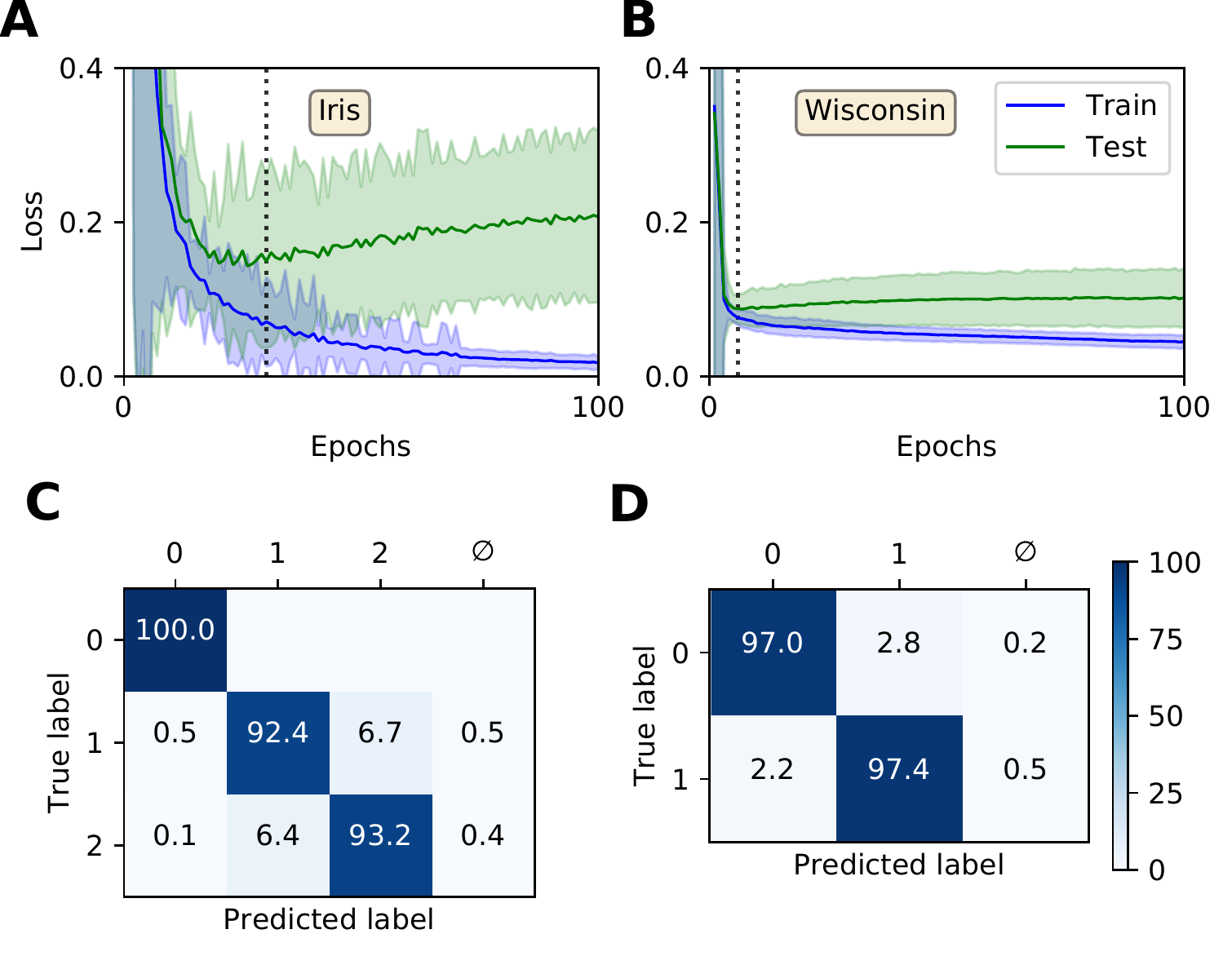}
\caption{
Training multilayer \acp{SNN} to classify the Iris and Wisconsin datasets using first-to-spike decoding. Iris and Wisconsin contain three and two pattern classes, respectively, with associated network structures: $48 \times 20 \times 3$ and $63 \times 20 \times 2$. (A--B): Evolution of the average training / test loss of the network on the two datasets, where the shaded region indicates the standard deviation. The vertical dashed line in each panel indicates the point at which early-stopping was applied, corresponding to the minimum recorded value in the test loss. Early-stopping was taken at 30 and 6 epochs for Iris and Wisconsin, respectively, which is equivalent to 30 and 24 (weight update) iterations. These results were obtained from 40 independent runs.
(C--D): Confusion matrices of the \acp{SNN}, post-training on Iris (C) and Wisconsin (D). The values report the percentage test accuracy, evaluated at the moment when early-stopping was applied. Iris data samples belonging to classes labelled `1' and `2' are linearly non-separable from each other, and for Wisconsin the labels `0' and `1' correspond to `benign' and `malignant', respectively. The null symbol, $\phi$, indicates that no clear prediction was formed by the network, being a consequence of either no output spiking or more than one output neuron sharing the same first spike response time (within the time resolution, $\delta t$).
}
\label{fig:iris_wisc_loss_confTe}
\end{figure}

For both Iris and Wisconsin the trained \acp{SNN} demonstrated success in fitting the data (Fig~\ref{fig:iris_wisc_loss_confTe}A--B), with final training accuracies of \num{99.88 \pm 0.04}{\%} and \num{98.04 \pm 0.04}{\%} after 100 epochs, respectively. In terms of their generalisation to test data, it was necessary to impose early-stopping to prevent overfitting. From multiple runs of the experiment, we determined the ideal training cut-off points to be approximately 30 and 6 epochs for Iris and Wisconsin, respectively. Since the number of weight update iterations / mini-batches per epoch was just one for Iris and four for Wisconsin, the equivalent number of iterations to early-stopping were 30 and 24, respectively. From the networks' confusion matrices (Fig~\ref{fig:iris_wisc_loss_confTe}C--D), evaluated at the point of early-stopping, the test accuracies were \num{95.2 \pm 0.2}{\%} and \num{97.12 \pm 0.08}{\%} for Iris and Wisconsin, respectively. As expected, the matrix for Iris indicated the relative challenge in separating the latter two, linearly non-separable classes. Furthermore, and as desired, the incidence of null predictions formed by the trained networks was kept to a minimum; in most cases a null prediction corresponded to a lack of firing activity in the network, typically preventing weight gradient computations, however this issue was mitigated by the addition of synaptic scaling in order to drive sufficient neuronal activation.

\begin{figure}[!tp]
\centering
\includegraphics{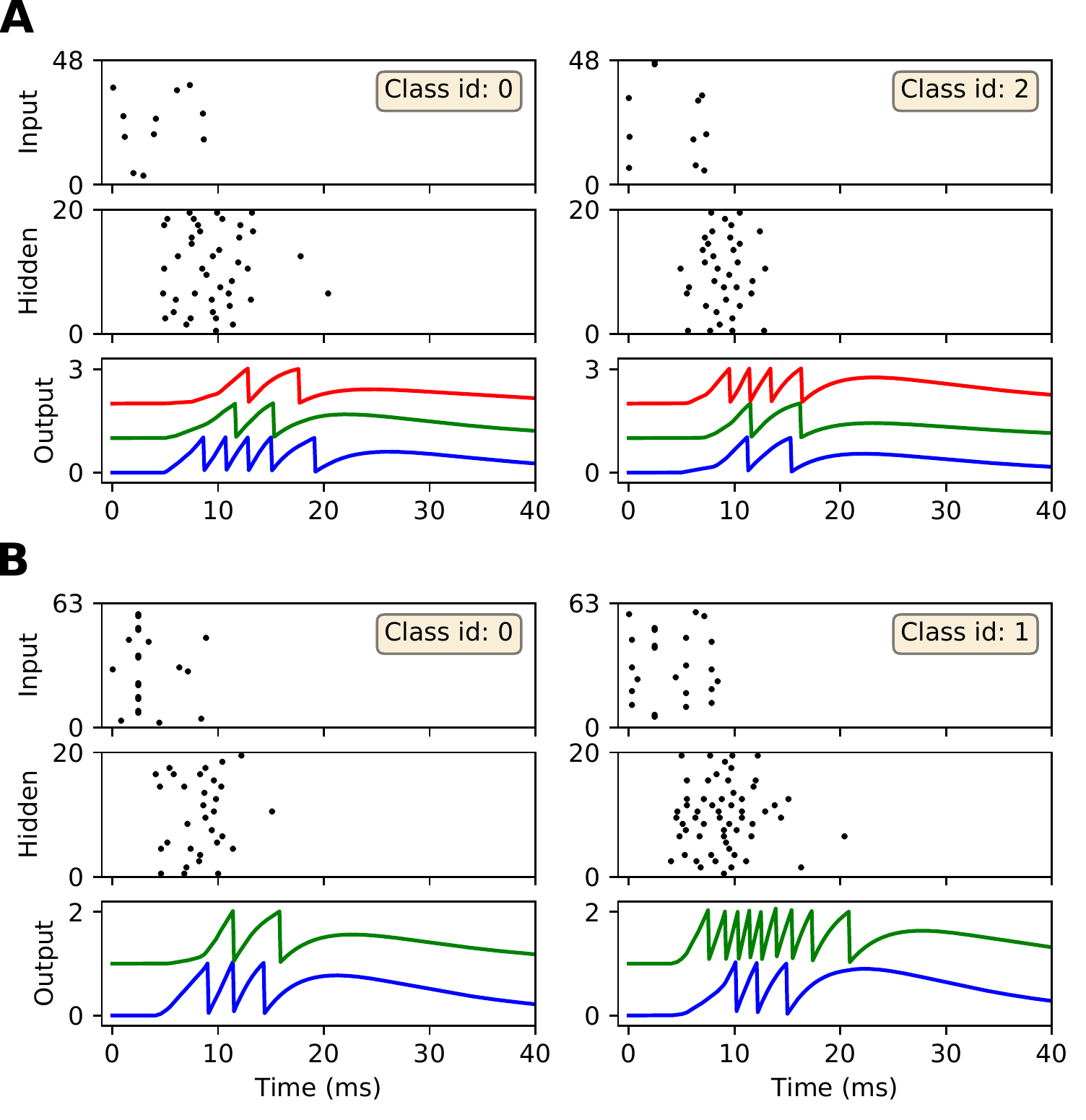}
\caption{
Spike rasters depicting network activity in response to selected Iris and Wisconsin data samples, corresponding to the experiment in Fig~\ref{fig:iris_wisc_loss_confTe} for networks trained with early-stopping applied. (A) The left and right main panels depict typical network responses to Iris samples belonging to the first and last classes, respectively. Both panels show spiking activity in the input, hidden and output layers of a trained network. (B) The left and right main panels depict network responses to benign (class id: 0) and malignant (class id: 1) Wisconsin samples, respectively. For both (A) and (B) desired first-spike responses in the output layer were observed, resulting in correct input classifications.
}
\label{fig:iris_wisc_rasters}
\end{figure}

The spiking activity of the \acp{SNN} in response to selected data samples, upon early-stopping during training on Iris and Wisconsin, respectively, is shown in Fig~\ref{fig:iris_wisc_rasters}. It is clear for each of the presented samples that input spikes were confined to the first \SI{10}{ms}, which prompted phasic activity in the hidden layer. With respect to the Iris samples (Fig~\ref{fig:iris_wisc_rasters}A), the network formed rapid, and correct first-spike reponses in the output layer: in this case within just \SI{10}{ms}. Due to the parameterisation of the learning rule, firing responses generated by the remaining neurons in the output layer were not completely eliminated: since these other neurons fired with sufficiently delayed onset, their resulting contribution to the output error signals used to inform weight updates were minimal. This behaviour was encouraged, given that it minimised data overfitting and prevented unstable dynamics arising due to competition between the backpropagation and synaptic scaling components of the learning rule (c.f. Eq~\eqref{eq:weight_update}). As with Iris, the network generated desired, rapid first-spike responses in the output layer when responding to Wisconsin data samples (Fig~\ref{fig:iris_wisc_rasters}B). In this example the Wisconsin-trained network formed correct predictions on the two selected samples, and interestingly a ramp-up in both hidden and output layer activity was observed for the malignant-labelled sample in order to shift the desired first-spike response earlier.

\subsection{Sensitivity to the Learning Schedule}

The previous experiments have demonstrated the performance of the first-to-spike classifier rule using optimal parameter selections of the learning coefficient $\eta_0$, used as part of the definition of RMSProp (Eq~\eqref{eq:RMSProp}). The sensitivity of the rule to its learning schedule is an important consideration regarding its versatility and application to unfamiliar data, therefore we tested its robustness when swept over a wide range of $\eta_0$ values.

As the test case we used the Iris dataset, with the same temporal encoding procedure and network setup as described in the previous section. As previously, stratified three-fold cross-validation was used to estimate the test loss during training. The regularisation parameter was set to $\lambda_0 = \num{E-3}$, and weights were constrained to values between $[-15, 15]$. Each epoch corresponded to one iterative weight update procedure, using a mini-batch size of 100. The network was trained for a total of 150 epochs for each $\eta_0$ selection, where at the end of each run we identified the minimum value in the recorded test loss and its associated number of epochs. The minimum number of epochs was determined by finding the first point at which the average test loss fell within \SI{1}{\%} of its lowest value, and its error was estimated based on the margin from falling within \SI{10}{\%} of the lowest value.

\begin{figure}[!tp]
\centering
\includegraphics{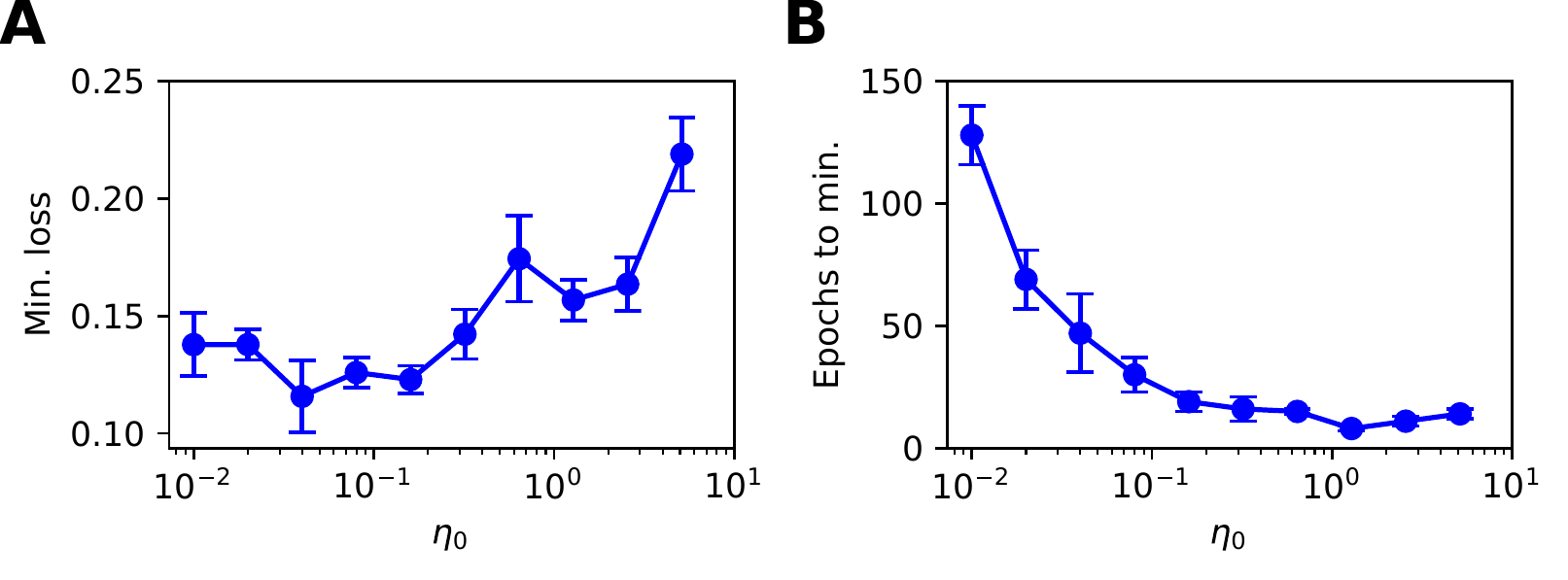}
\caption{
Parameter sweep over the RMSProp learning coefficient $\eta_0$, evaluated for an \ac{SNN} trained on Iris with the same network structure and training procedure as in Fig~\ref{fig:iris_wisc_loss_confTe}. (A) Minimum recorded test loss as a function of $\eta_0$, taken as the median value, where error bars indicate the \ac{SEM}. (B) The associated number of training epochs needed to reach the minimum test loss. These results were gathered from 20 independent runs.
}
\label{fig:iris_minima}
\end{figure}

The minimum test loss attained, including the associated number of training epochs, is shown in Fig~\ref{fig:iris_minima} for selections of $\eta_0$ between \num{E-2} and \num{E1}. From these results, it follows that a learning coefficient with a value of around \num{E-1} provided a reasonable trade-off between network performance and learning speed: larger $\eta_0$ values returned sub-optimal test loss minima, while smaller values led to an exponential increase in the training time with little change in the performance. Extended parameter sweeps also indicated similar behaviour with respect to the Wisconsin dataset as the test case. For these reasons we were motivated to select an optimal value of $\eta_0 = \num{E-1}$ for both Iris and Wisconsin.

These observations support our selection of RMSProp, as opposed to a fixed learning rate which demonstrated greater sensitivity to its parameter choice. In addition to the above, we found that RMSProp's learning coefficient exhibited a dependence on the number of input synapses per network layer, such that more optimal performance was attained by adjusting $\eta_0$ proportional to $1 / N_l$. This was used to inform our choice of $\eta_0$ used for the MNIST dataset.

\subsection{Classifying MNIST: Latency Encoding}

The MNIST dataset of handwritten digits \citep{LeCun1998} is commonly used to benchmark a classifier system's performance, owing to the relatively high structural complexity of the data and the large number of training and test samples. Although this problem is largely solved using deep convolutional neural networks, MNIST still poses a challenge to solve using spike-based computational approaches. For this reason, we apply our first-to-spike classifier rule to solving MNIST in order to get an indication of its potential for real world data classification.

MNIST consists of \num{60}{k} and \num{10}{k} training and test samples, respectively, where each sample corresponds to a \num{28x28}, 8-bit grayscale image depicting a handwritten digit between 0 and 9. In order to make these real-valued images compatible with spike-based processing we applied a latency encoding strategy: forming a one-one association between each input pixel and an encoding \ac{LIF} neuron. In this way, each image, consisting of 784 pixels, was transformed into 784 single-spike latencies presented by the first layer of a multilayer \ac{SNN}. Specifically, and as described in section~\ref{subsec:temporal_encoding}, the pixel values were transformed into current intensities using the scaling factor $I_\mathrm{max} = \SI{20}{nA}$, resulting in the following pixel-to-latency mapping: $[84, 256) \mapsto [9, 2) \mathrm{ms}$, where pixel values less than 84 were insufficient to elicit an encoded response. In terms of network structure, the simulated \acp{SNN} consisted of $784 \times N_{2} \times 10$ neurons, where the number of hidden neurons, $N_2$, was varied, and the the number of output neurons was matched to the 10 digit classes. According to the first-to-spike decoding strategy, the first output neuron to respond with a spike predicted the input sample's class. The network weights were initialised by drawing hidden and output values from uniform distributions over the ranges: $[0, 0.4)$ and $[0, 32 / N_2)$, respectively. The softmax scale parameter was set to $\nu = 4$ in order to tighten the conditional probability distribution of class label predictions formed by the output layer: this choice was supported by preliminary simulations, where it was found to boost the discriminative power of the \ac{SNN} when handling a larger number of input classes. In terms of network training, at the start of each run 600 of the MNIST training samples were set aside in a stratified manner for validation. The remaining training samples were then iteratively presented to the network as mini-batches, with a total of 4000 iterations. To get an indication of the network's performance during training the loss on the validation data was computed every 20 iterations. The regularisation parameter and RMSProp coefficient were set to $\lambda_0 = \num{E-4}$ and $\eta_0 = 0.01$, respectively. Throughout training, all weights were constrained to values in the range $[-2, 2]$ to avoid overfitting.

\begin{figure}[!tp]
\centering
\includegraphics{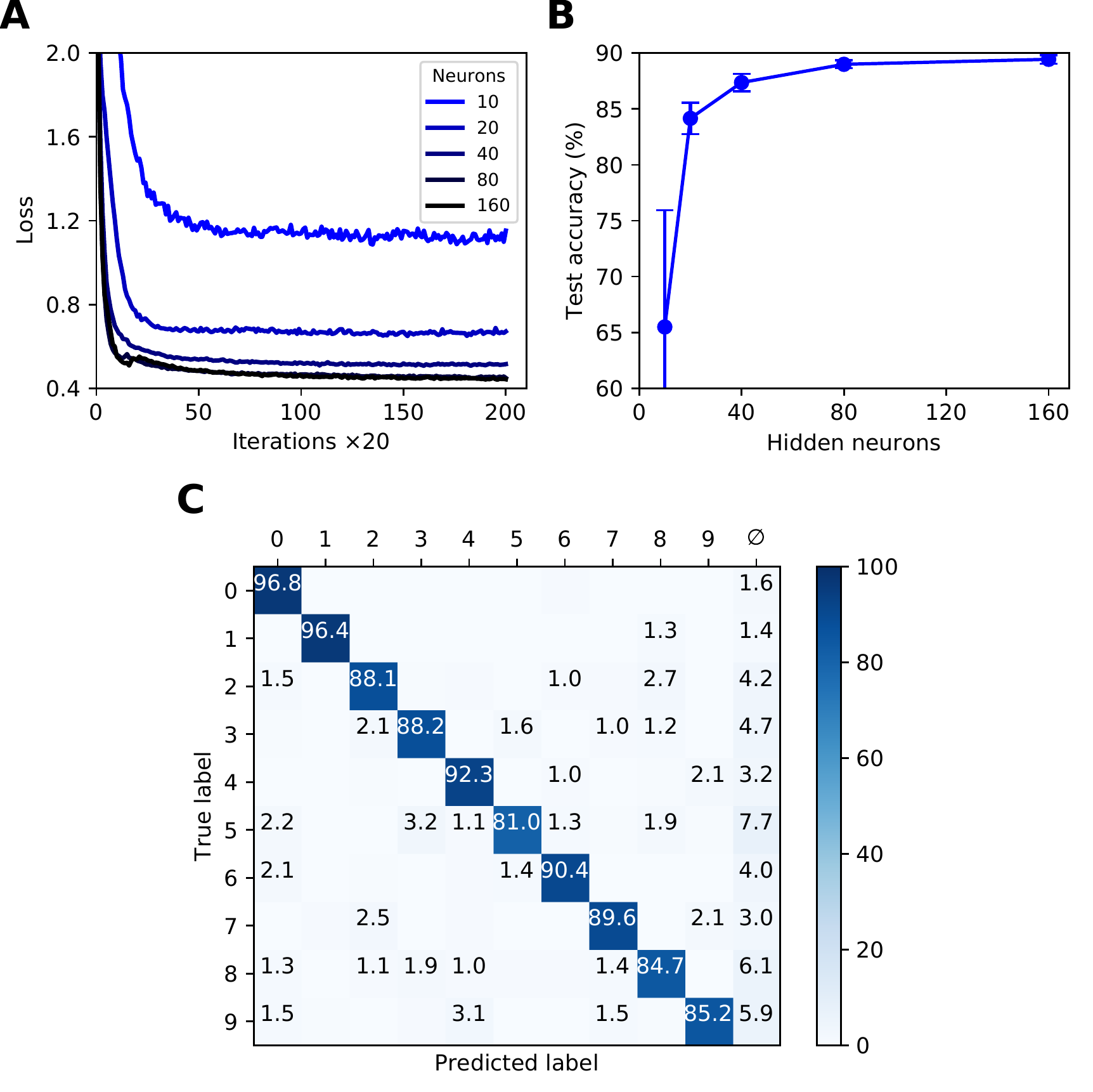}
\caption{
Multilayer \acp{SNN} trained on the MNIST dataset, using a latency encoding strategy to transform pixels into single, precisely-timed input spikes. (A) Network loss as a function of the number of training iterations, evaluated using a validation set. Each curve corresponds to a different number of hidden layer neurons, $N_2$, from between 10 (lightest colour) and 160 (darkest colour). (B) The final accuracy of the network after 4000 training iterations, as evaluated on the \num{10}{k} test samples. (C) Confusion matrix of an \ac{SNN} containing 160 hidden neurons, as evaluated on the test samples at the end of training. Values less than \SI{1}{\%} are not indicated. These results were averaged from 10 independent runs.
}
\label{fig:mnist_latency}
\end{figure}

As shown by Fig~\ref{fig:mnist_latency}, the trained \acp{SNN} were capable of generalising from the MNIST training samples to the withheld test samples, with a highest recorded test accuracy of \num{89.4 \pm 0.4}{\%} for a network containing 160 hidden layer neurons. With the given selection of regularisation parameters and weight constraints, model overfitting was minimised and smooth convergence was observed within the maximum number of training iterations (Fig~\ref{fig:mnist_latency}A). Moreover, as the hidden layer size was increased, a speedup in the learning rate became apparent. As indicated by Fig~\ref{fig:mnist_latency}B, the accuracy of the network approached an asymptotic value of just under \SI{90}{\%} when containing up to 160 neurons. The confusion matrix depicted in Fig~\ref{fig:mnist_latency}C corresponds to a network containing 160 neurons, and provides some insight into the robustness of network classifications with respect to each of the presented MNIST test digits. As expected, digits `zero' and `one' were least challenging for the network to identify by virtue of their distinctive features, whereas the digit `five', for example, tended to share a greater feature overlap with several other digits, making it somewhat more difficult to discriminate. Furthermore, in the event of a digit not being recognised by the network there was a tendency towards a null prediction (no output spikes) being returned rather than an erroneous digit; technically this is a preferable outcome, since it reduces the likelihood of false-positives by the network.

In summary, the first-to-spike classifier rule has demonstrated generalisation capability on the MNIST dataset to a reasonable degree of accuracy. The results reported here do not reflect an upper bound on the classifier's MNIST performance, and with further parameter tuning and feature preprocessing further accuracy gains would be expected. For simplicity, this experiment considered a straightforward one-one mapping between each input pixel and encoding neuron in order to transform the data, although such a scheme is computationally prohibitive for spike-based processing and fails to fully exploit the precise timings of individual spikes. Utilising a fully temporal encoding strategy presents the next challenge, and is addressed in the following section.

\subsection{Classifying MNIST: Scanline Encoding}

So far, the technical capability of the first-to-spike classifier rule has been demonstrated on MNIST when encoded using a one-one association between input pixels and single-spike latencies. This encoding strategy is somewhat simplistic, however, and fails to take full advantage of the precise timings of multiple spikes as a means to perform dimensionality reduction on the data. To address this we consider an alternative encoding strategy, termed scanline encoding, that extends on the work of \cite{Lin2018}, and enables more compact feature representations of the MNIST digits using substantially fewer encoding neurons.

In order to transform the real-valued features of MNIST into sequences of precisely-timed spikes, we applied the scanline encoding strategy as described in \ref{subsec:temporal_encoding}. In summary, at the start of each simulation run we implemented a variable number of scanlines, $n_s$, ranging between 8 and 64. Each of these scanlines had an arbitrarily determined orientation, independent of the others, and was constrained to intersect through a point close to the centre of the image space. For the duration of each run these scanlines were held fixed, and in response to each presented sample a scanline read-in its sequence of input pixels and returned a time-varying current; this current in turn acted as the stimulus for an encoding \ac{LIF} neuron in the first layer of an \ac{SNN}, driving a spike train response (Fig~\ref{fig:scanline_encoding} illustrates this scanning process for an example image). Hence, the strategy we employed here was capable of transforming high-dimensional images into compact spatiotemporal spike patterns, whilst still retaining characteristics of their spatial organisation. With respect to the network structure, \acp{SNN} consisting of $N_1 \times N_2 \times 10$ neurons were simulated, where the number of input neurons was matched to the number of scanlines used, $N_1 = n_s$, and the number of hidden neurons $N_2$ was varied. As previously, the number of output neurons was matched to the 10 different digit classes of MNIST, and first-to-spike decoding was used to classify the data samples. In terms of network connections, two distinct modelling approaches were considered regarding input-to-hidden layer spike propagation: `delayless' and `delayed'. In the delayless case, spikes were instantaneously transmitted from input to hidden neurons, as has been implemented so far for all the previous experiments. In the delayed case, however, spikes transmitted from input to hidden neurons were subject to propagation delay: ranging from between \num{1} and \SI{10}{ms}, rounded to the nearest millisecond. At the start of each experimental run, these propagation, or conduction, delays were randomly drawn from a uniform distribution for each input-to-hidden connection, and held fixed thereafter. In all cases hidden-to-output layer spike conduction was delayless. The purpose of simulating conduction delays was to determine if this could assist a network in linking early / late spike arrival times arising from scanline-encoded digits. With respect to weight initialisation, hidden and output weights were initialised according to uniform distributions with values ranging between $[0, 40 / N_1)$ and $[0, 32 / N_2)$, respectively. The softmax scale parameter was set to $\nu = 4$. For each run, a network was trained and validated in a similar way to the latency encoding experiment: 600 MNIST training samples were set aside for validation every $20^\mathrm{th}$ iteration, and the remaining samples were iteratively presented as mini-batches for 1600 iterations. The regularisation parameter and RMSProp coefficient were set to $\lambda_0 = \num{E-4}$ and $\eta_0 = 0.05$, respectively. All of the network weights were constrained to values in the range $[-6, 6]$.

\begin{figure}[!tp]
\centering
\includegraphics{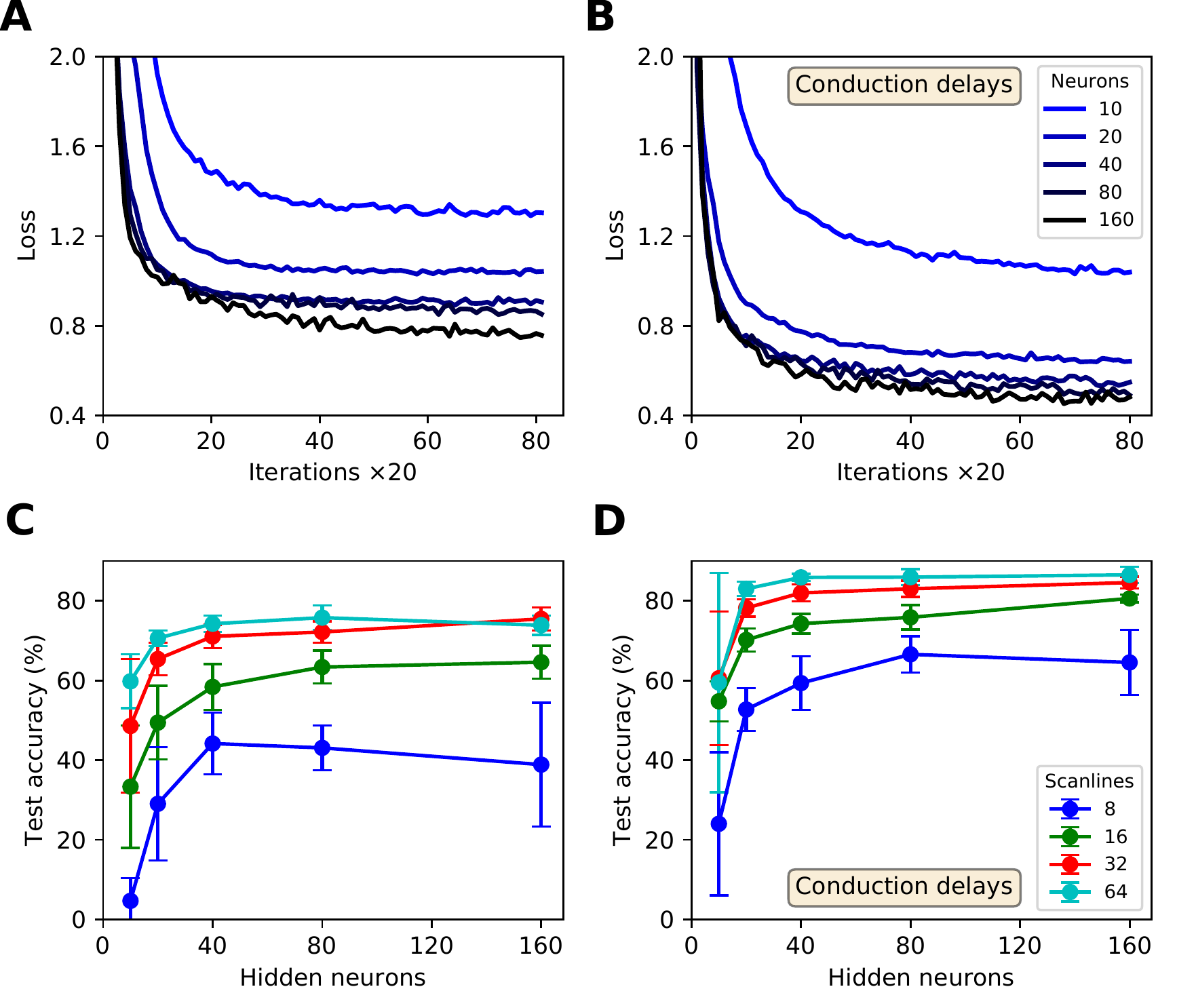}
\caption{
Multilayer \acp{SNN}, with or without input-to-hidden layer spike-propagation (conduction) delays, trained on the MNIST dataset. Scanline encoding was used to transform images into compact spatiotemporal spike patterns. (A and B) Network losses for 32 scanlines as a function of the number of training iterations, evaluated using a validation set. Panels A and B correspond to `delayless' and `delayed' networks, respectively: referring to the inclusion of conduction delays between the input and hidden layers. For each panel, the curve colour corresponds to a certain number of hidden layer neurons, $N_2$: from between 10 (lightest colour) and 160 (darkest colour). (C and D) The final accuracies of the network after 1600 training iterations, as evaluated on the \num{10}{k} test samples. Panels C and D correspond to delayless and delayed networks, respectively. These results were averaged from 10 independent runs.
}
\label{fig:mnist_scan_results}
\end{figure}

Shown in Fig~\ref{fig:mnist_scan_results} are results obtained from trained \acp{SNN}, with and without conduction delays between the input and hidden layers, for scanline-encoded MNIST samples. It can be seen that in all cases the trained networks converged in learning within the maximum 1600 iterations, with an additional small speedup as the number of hidden neurons was increased from 10 to 160 (Fig~\ref{fig:mnist_scan_results}A-B). It is also apparent that the inclusion of conduction delays reduced the final loss values, enabling the network to approach the same values as reported for latency-encoded MNIST (compare with Fig~\ref{fig:mnist_latency}A). In terms of the post-training performance on the withheld test samples, the highest attained accuracies were \num{76 \pm 6}{\%} and \num{87 \pm 2}{\%} for delayless and delayed networks, respectively, as evaluated using 160 hidden neurons and between 32 and 64 scanlines (Fig~\ref{fig:mnist_scan_results}C-D). From the gathered results, there was diminishing returns in the final test accuracy as the number of neurons and scanlines were increased beyond 160 and 32, respectively. Comparatively, the test accuracy returned by the best performing network with conduction delays fell within just \num{3}{\%} of that obtained based on a latency encoding strategy, but with the advantage of using an order of magnitude fewer input neurons.

\begin{figure}[!tp]
\centering
\includegraphics{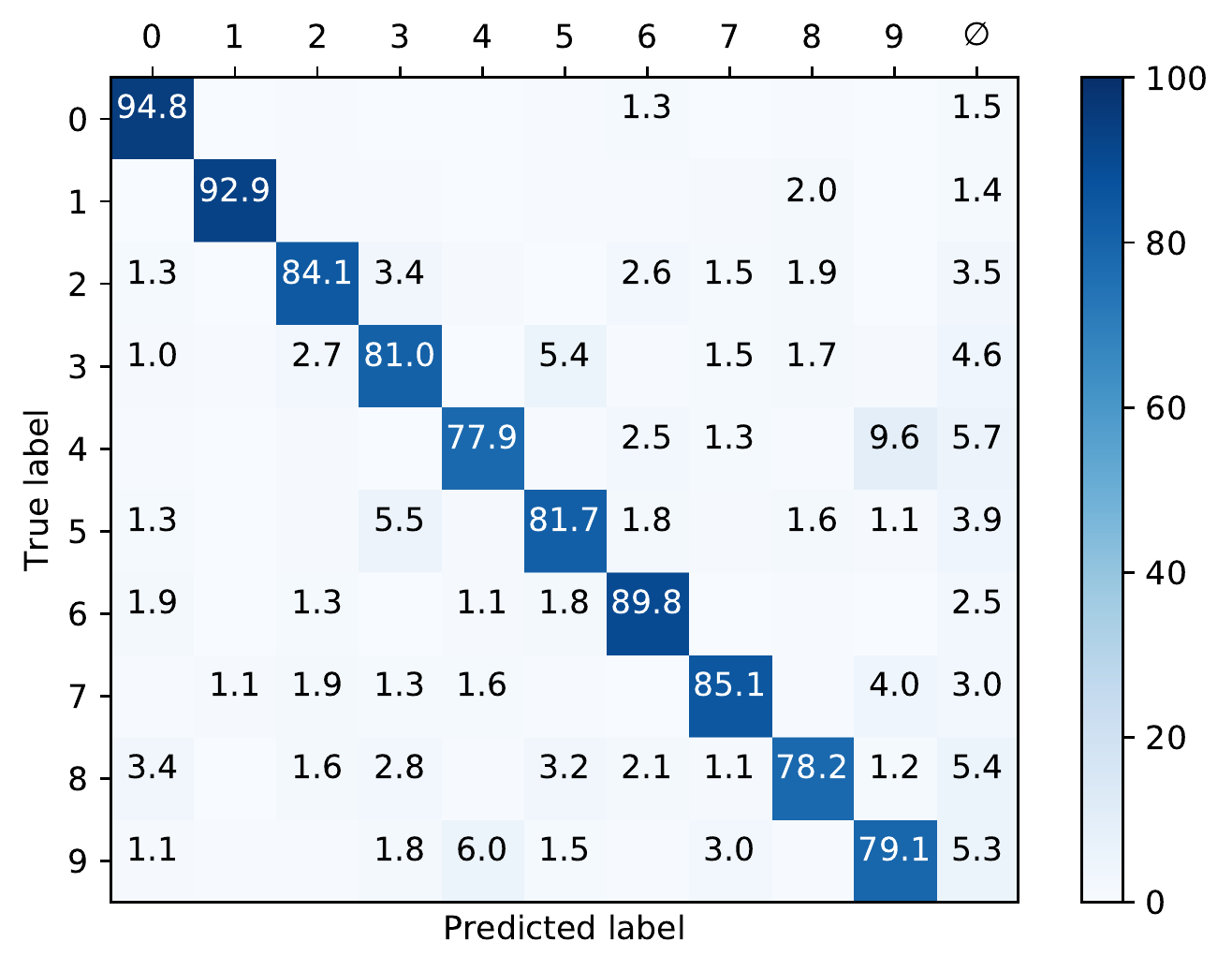}
\caption{
Confusion matrix of an \acp{SNN} with input-to-hidden layer conduction delays, after 1600 training iteration on MNIST encoded using 32 scanlines. This was evaluated on the withheld test samples. The network contained 160 hidden neurons. Values less than \SI{1}{\%} are not indicated. These results were averaged from 10 independent runs.
}
\label{fig:mnist_scan_confusion_delay}
\end{figure}

The confusion matrix of a post-trained $32 \times 160 \times 10$ network with conduction delays, as evaluated on the withheld MNIST test samples, is depicted in Fig~\ref{fig:mnist_scan_confusion_delay}. These results correspond to a high performing case from Fig~\ref{fig:mnist_scan_results}. As expected, the network demonstrated the least difficulty in recognising the digits `zero' and `one', closely followed by `six'. However, the network tended to confuse the digits `four' and `nine' with relatively high frequency, owing to their closer similarities. By comparison with the confusion matrix from Fig~\ref{fig:mnist_latency}C, the overall percentage of null predictions by the networks for both encoding strategies were consistent. Despite this, networks utilising scanline encoding gave rise to more variable predictions between experiment runs: the coefficients of variation with respect to correct predictions was \num{0.08 \pm 0.03} and \num{0.02 \pm 0.01} for scanline- and latency-based encoding, respectively; this discrepancy was attributed to the random selection of scanline orientations between runs, whereas latency-based representations of the same input samples remained fixed.

\begin{figure}[!tp]
\centering
\includegraphics[width=\textwidth]{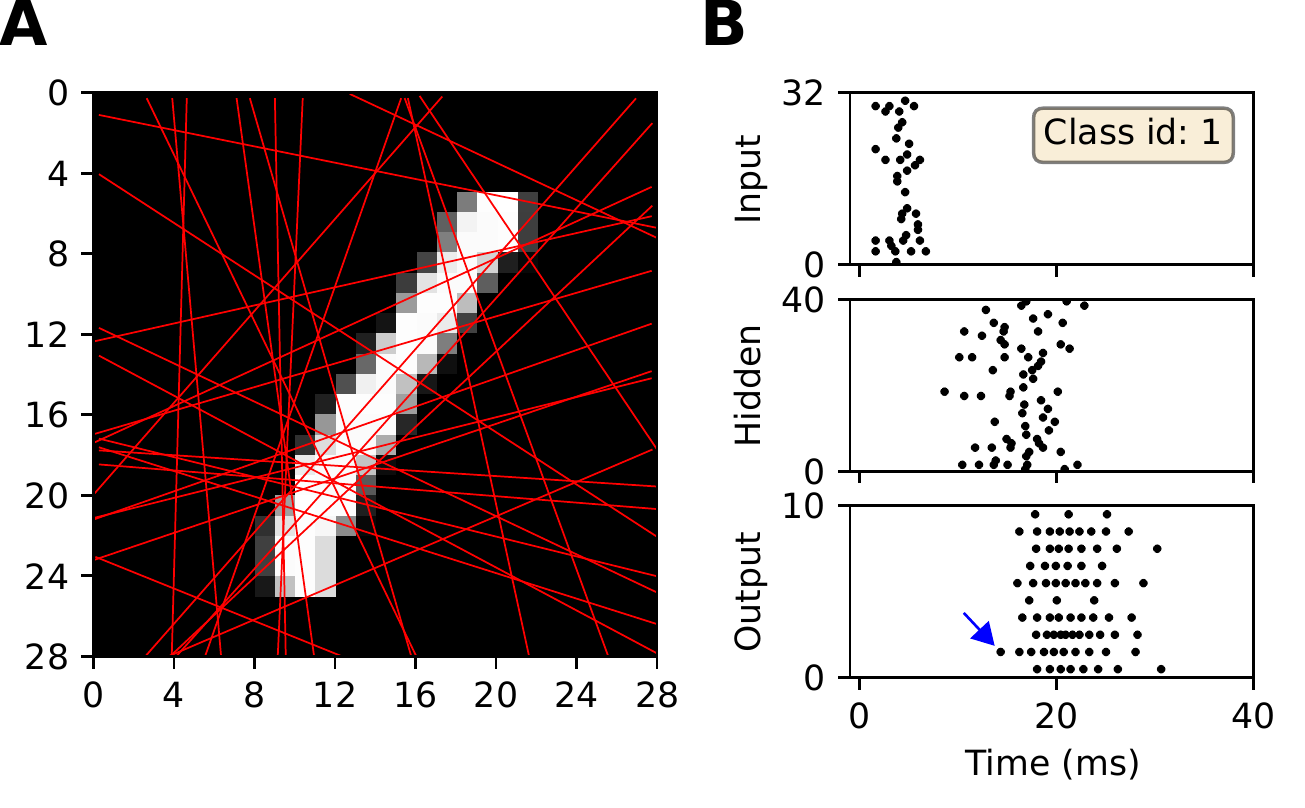}
\caption{
Illustration of the encoding and subsequent classification of the MNIST digit `one' based on scanlines. In this example, the digit is first transformed into spike trains via scanline encoding, before being processed by a previously trained \ac{SNN} containing 40 hidden neurons and input-to-hidden layer conduction delays. (A) The 32 scanlines encoding the digit `one' (red lines). (B) Spike raster of the network's response to the encoded digit. The top, middle and bottom subplots correspond to input, hidden and output spike patterns, respectively. In this case, the first neuron to respond with a spike (indicated by the blue arrow) corresponds to the desired class label, resulting in a correct classification.
}
\label{fig:mnist_scan_raster_delay}
\end{figure}

For illustrative purposes, an example of a scanline-encoded digit and the response it elicited in a post-trained \ac{SNN} is shown in Fig~\ref{fig:mnist_scan_raster_delay}. A sample of the digit `one', which was withheld during network training, was transformed into sequences of precisely-timed spikes, and represented by the first layer of neurons in a minimal $32 \times 40 \times 10$ network with input-to-hidden layer conduction delays. In this example, the network correctly identified the input sample by driving the corresponding output neuron to respond first with a spike. As indicated by panel A, most of the feature space was covered by the 32 scanlines: an increase in this number resulted in diminishing returns, relating to feature redundancy. In terms of the spike raster subplots in panel B, there is a relatively large offset in the emergence of hidden spiking with respect to the onset of input spikes, whereas there is a large degree of overlap between hidden-and-output spiking. This reflects the delayed propagation of input-to-hidden spikes, but which affords the network time to link early and late input spikes in order to inform its final decision. The activity of neurons in the output layer tended to be greater than that observed with the other datasets (compare with Fig~\ref{fig:iris_wisc_rasters}), indicative of the increased complexity in learning to discriminate between a larger number of classes with more feature overlap.

To summarise, this section has demonstrated a novel methodology to training \acp{SNN} on MNIST that are more constrained in their size, and yet more efficient in terms of their spike-based processing. This has been realised by the application of scanline encoding: a feature preprocessing method that can transform high dimensional images into compact spatiotemporal spike patterns. In particular, the results obtained using this method approach a similar level of accuracy to networks trained via a more commonly used, but more computationally expensive, one-one encoding scheme. Relying more on the precise timings of individual spikes for data classification massively reduces the number of encoding neurons required, and could find important applications in constrained network architectures, for example in neuromorphic systems like Heidelberg's HICANN-DLS device \citep{Friedmann2017}. It is expected that the performance of this method could be improved upon by making scanline encoding more domain specific: for example by optimising scanline orientations prior to network training, rather than setting them arbitrarily. Our choice of a random initialisation, however, indicates the potential in transferring this method to unfamiliar problem domains.

\section{Discussion}

In this article we have introduced a new supervised approach to training multilayer spiking neural networks, using a first-to-spike decoding strategy, with the proposed learning rule capable of providing robust performance on several benchmark classification datasets. The learning rule extends on our previous formulation in \cite{Gardner2015} by redefining the network's cost function to depend on the distribution of first spike arrival times in the output layer, rather than entire spike trains, and specifying the target signal according to which one of $c$ output neurons should be driven to fire first; this redefinition of the cost function is particularly advantageous for data classification purposes since it places much less of a constraint on the network's parameters during training, thereby avoiding overfitting of the data. Furthermore, restricting our focus to just first-spike arrival times in the output layer has allowed us to largely reduce the runtime of simulations: an important consideration when taking into account the relatively high computational cost in simulating spike-based models. Based on first-to-spike decoding, pattern classification was rapid: with predictions in some cases made within just \SI{10}{ms} of pattern onset. Such a decoding strategy has similarly been used to good effect in \citep{Mostafa2017,Bagheri2018}, and moreover avoids the ambiguity of decision making based on comparisons between entire target and actual output spike trains as used in \citep{Bohte2002a,Florian2012,Sporea2013,Gardner2015,Gardner2016}. We highlight the novel, hybrid nature of our learning model: which implements both deterministic, \ac{LIF}-type output neurons for more reliable network responses, and stochastic hidden layer neurons that should aid with its regularisation.

The formulation of our learning rule combines several different techniques, as found in \cite{Bohte2002a, Pfister2006, Gardner2015, Mostafa2017}. As our first step, we selected the network's cost function as the cross-entropy, dependent on the distribution of first-spike arrival times in the output layer, and set the target signal according to the index of the neuron associated with the correct class \citep{Mostafa2017}. Subsequently, and in order to apply the technique of spike-based backpropagation, we estimated the gradients of deterministically-generated output firing times by following the linear approximation used in \cite{Bohte2002a}; our choice here was motivated by simplicity reasons, since only single, first-spikes in the output layer were required for parameter optimisation. We then applied our previously described probabilistic method to solving hidden layer spike gradients for stochastic neurons \citep{Gardner2015}, which supports multiple firing times and is theoretically justified \citep{Pfister2006}. Therefore, in combining these methods, we arrive at a fully-fledged, supervised learning rule for multilayer \acp{SNN} that is extensible to $N$ number of hidden layers.

In terms of performance, we first demonstrated the learning rule to be capable of solving the non-trivial \ac{XOR} classification task: establishing its ability to classify linearly non-separable data, for which a hidden layer is required. Subsequently, the rule was found to be highly accurate in classifying more challenging data samples belonging to the Iris and Wisconsin datasets, and for which two of the Iris classes are linearly non-separable. In particular, we found that implementing the RMSProp learning schedule \citep{Hinton2012} made the network less sensitive to the choice of learning coefficient $\eta_0$, suggesting it as an effective mechanism to minimising the process of parameter fine-tuning; this in turn increased the flexibility of the rule as applied to different datasets. Additionally, we found in general that regularising the network by penalising high neuronal firing rates resulted in improved generalisation ability and accuracy, confirming the observations of \cite{Zenke2018}. Suppressing high firing rates also reduced the number of computational operations in the run simulations, since fewer hidden spikes were integrated over when computing the iterative weight updates, thereby enabling a large speedup in runtime. With respect to the rule's performance on MNIST, the test accuracy didn't reach state-of-the art: with our rule achieving around \SI{90}{\%}. It is noted, however, that attaining high accuracy on MNIST, including other structurally complex datasets such as Fashion-MNIST and ImageNET \citep{Xiao2017,Deng2009}, via spike-based processing still poses more of a challenge compared with traditional \ac{ANN} approaches. Many existing spike-based supervised learning methods have achieved accuracies ranging between \SI{90} and almost \SI{99}{\%} on MNIST, with the highest levels relating to deeper \ac{SNN} architectures and convolutional filtering optimised for image processing tasks \citep{Connor2013,Neftci2014,Diehl2015,Lee2016,Mostafa2017,Tavanaei2019,Zenke2020}. In our approach, we tested minimal \ac{SNN} architectures with their flexibility and transferrability to constrained neuromorphic hardware platforms such as HICANN-DLS \citep{Friedmann2017} in mind. Intriguingly, however, we found that our rule was still capable of achieving almost \SI{90}{\%} accuracy on MNIST when using our novel scanline encoding method, as inspired by \cite{Lin2018}, and relying on as little as 32 encoding input neurons. We also note that these results represent a lower bound on what is achievable using our learning method: with model refinement in the form of extended hyper-parameter fine-tuning and deeper network architectures, it is expected that the final accuracies could be further increased.

\section*{Appendix: Extending the Learning Rule}

\begin{figure}[!tp]
\centering
\includegraphics{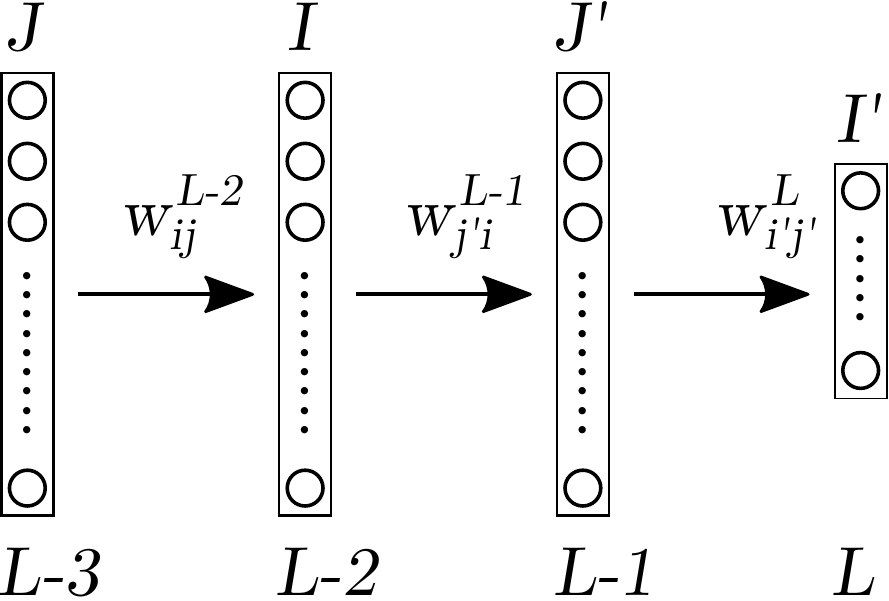}
\caption{
Schematic of the multilayer \ac{SNN} architecture considered here, containing two hidden layers. The network is feedforward with all-to-all connectivity between adjacent layers. From the last layer, $L$, to the first layer, $L-3$, neurons are respectively indexed: $i' \in I'$, $j' \in J'$, $i \in I$, and $j \in J$. Synaptic weights are denoted with respect to the postsynaptic layer, for example $w_{ij}^{L-2}$ is the weight projecting from presynaptic neuron $j$ onto postsynaptic neuron $i$ residing in layer $L-2$.
}
\label{fig:snn_schematic}
\end{figure}

The procedure used to derive weight updates in multilayer \acp{SNN} is also extensible to deeper network architectures. For demonstrative purposes, we consider a multilayer \ac{SNN} containing two hidden layers, and derive weight updates for hidden neurons residing in the third-last layer, i.e. for layer $l = L - 2$; refer to Fig~\ref{fig:snn_schematic} for a schematic of this network structure, and for the notation we use from this point onwards.

\paragraph{Weight update rule.} Gradient descent is applied to Eq~\eqref{eq:cross-entropy-onehot} with respect to weights projecting onto layer $L - 2$, according to
\begin{equation} \label{eq:grad_wh2}
\Delta w_{ij}^{L-2} = -\eta \frac{\partial C(\mathbf{y}, \mathbf{a}^L)}{\partial w_{ij}^{L-2}} \;.
\end{equation}
Similarly as in Eq~\eqref{eq:grad_wh_2}, the gradient term is expanded as follows:
\begin{align} \label{eq:grad_wh2_2}
\frac{\partial C(\mathbf{y}, \mathbf{a}^L)}{\partial w_{ij}^{L-2}}
	&= \sum_{i' \in I'} \frac{\partial C(\mathbf{y}, \mathbf{a}^L)}{\partial u_{i'}^L} \frac{\partial u_{i'}^L(t)}{\partial w_{ij}^{L-2}} \bigg\rvert_{t = \tau_{i'}} \nonumber \\
	&= \sum_{i' \in I'} \delta_{i'}^L \frac{\partial u_{i'}^L(t)}{\partial w_{ij}^{L-2}} \bigg\rvert_{t = \tau_{i'}} \;,
\end{align}
where for convenience we use the identity of the output error signal, defined by Eq~\eqref{eq:errorL}. Hence, using the \ac{SRM} defined by Eq~\eqref{eq:potential}, the gradient of the ${i'}^\mathrm{th}$ output neuron's membrane potential is given by
\begin{align} \label{eq:grad_wh2_3}
\frac{\partial u_{i'}^L(t)}{\partial w_{ij}^{L-2}}\bigg\rvert_{t = \tau_{i'}}
&= \sum_{j' \in \Gamma_{i'}^L} w_{i'j'}^L \frac{\partial}{\partial w_{ij}^{L-2}} \left( \epsilon \ast S_{j'}^{L-1} \right)(\tau_{i'}) \nonumber \\
&= \sum_{j' \in \Gamma_{i'}^L} w_{i'j'}^L \left( \epsilon \ast \frac{\partial S_{j'}^{L-1}}{\partial w_{ij}^{L-2}} \right)(\tau_{i'}) \;,
\end{align}
where $\Gamma_{i'}^L$ is the set of immediate neural predecessors of a last layer neuron $i'$. As in section~\ref{subsec:learning_rule}, we substitute the ill-defined gradient of a hidden layer spike train with the gradient of its expected value, and consider spikes to be distributed according to an underlying, instantaneous firing rate: $\rho_{j'}^{L-1}(t) = g(u_{j'}^{L-1}(t))$. Hence, following the same approach as Eqs~\eqref{eq:grad_wh_4} and \eqref{eq:grad_wh_5}, we arrive at the following expression for the gradient of the expected hidden spike train:
\begin{equation} \label{eq:grad_wh2_4}
\frac{\partial \left\langle S_{j'}^{L-1}(t) \right\rangle_{S_{j'}^{L-1}|L-2, j'}}{\partial w_{ij}^{L-2}}
= \delta_D(t-\hat{t}) \frac{\partial \rho_{j'}^{L-1}(t|L-2, j')}{\partial w_{ij}^{L-2}} \;,
\end{equation}
where the expected spike train is conditioned on activity in the previous layer, $L-2$, and its own last firing time, $\hat{t}_{j'}$. As before, $\delta_D$ is the Dirac-delta function: depending on an arbitrary, last firing time, $\hat{t}$. Combining the above with Eqs~\eqref{eq:potential} and \eqref{eq:escape_rate}, we find
\begin{align} \label{eq:grad_wh2_5}
\frac{\partial \left\langle S_{j'}^{L-1}(t) \right\rangle_{S_{j'}^{L-1}|L-2, j'}}{\partial w_{ij}^{L-2}}
&= \frac{1}{\Delta u} \delta_D(t-\hat{t}) \rho_{j'}^{L-1}(t|L-2, j') \frac{\partial u_{j'}^{L-1}(t)}{\partial w_{ij}^{L-2}} \nonumber \\
&= \frac{1}{\Delta u} \left\langle S_{j'}^{L-1}(t) \frac{\partial u_{j'}^{L-1}(t)}{\partial w_{ij}^{L-2}} \right\rangle_{S_{j'}^{L-1}|L-2, j'} \;.
\end{align}
Using Eq~\eqref{eq:potential}, the gradient of the $j^{'\mathrm{th}}$ neuron's membrane potential is expanded with respect to the previous layer as follows:
\begin{align} \label{eq:grad_wh2_6}
\frac{\partial u_{j'}^{L-1}(t)}{\partial w_{ij}^{L-2}}
&= w_{j'i}^{L-1} \frac{\partial}{\partial w_{ij}^{L-2}} \left( \epsilon \ast S_i^{L-2} \right)(t) \nonumber \\
&= w_{j'i}^{L-1} \left( \epsilon \ast \frac{\partial S_i^{L-2}}{\partial w_{ij}^{L-2}} \right)(t) \;,
\end{align}
where all spike train gradients in layer $L-2$ except for $i^\mathrm{th}$ one do not depend on $w_{ij}^{L-2}$, and therefore vanish. As usual, the gradient of a hidden spike train is substituted with the gradient of its expected value, conditioned on the neuron's last firing time and its received spike trains from the previous layer:
\begin{align} \label{eq:grad_wh2_7}
\frac{\partial \left\langle S_{i}^{L-2}(t) \right\rangle_{S_i^{L-2}|L-3, i}}{\partial w_{ij}^{L-2}}
&= \delta_D(t-\hat{t}) \frac{\partial \rho_i^{L-2}(t|L-3, i)}{\partial w_{ij}^{L-2}} \nonumber \\
&= \frac{1}{\Delta u} \left\langle S_i^{L-2}(t) \frac{\partial u_i^{L-2}(t)}{\partial w_{ij}^{L-2}} \right\rangle_{S_i^{L-2}|L-3, i} \;,
\end{align}
where hidden spikes in layer $L-2$ are distributed according to $\rho_i^{L-2}$.
The gradient of the neuron's membrane potential is determined using Eq~\eqref{eq:potential}:
\begin{equation} \label{eq:grad_wh2_8}
\frac{\partial u_i^{L-2}(t)}{\partial w_{ij}^{L-2}} = \left( \epsilon \ast S_j^{L-3} \right)(t) \;.
\end{equation}
Hence, combining Eqs~\eqref{eq:grad_wh2_5} to \eqref{eq:grad_wh2_8} provides the gradient of the expected value of the hidden spike train in layer $L-1$:
\begin{align} \label{eq:grad_wh2_9}
&\frac{\partial \left\langle S_{j'}^{L-1}(t) \right\rangle_{S_{j'}^{L-1}|L-2, j'}}{\partial w_{ij}^{L-2}} \nonumber \\
&\quad= \frac{w_{j'i}^{L-1}}{\left(\Delta u\right)^2} \left\langle S_{j'}^{L-1}(t) \left( \epsilon \ast \left\langle S_i^{L-2} \left( \epsilon \ast S_j^{L-3} \right) \right\rangle_{S_i^{L-2}|L-3, i} \right)(t) \right\rangle_{S_{j'}^{L-1}|L-2, j'} \;.
\end{align}
As usual, we estimate the above expected gradient by sampling hidden spike trains on each simulation run. Therefore, dropping the expectation operators provides the gradient estimator:
\begin{equation} \label{eq:grad_wh2_10}
\frac{\partial \left\langle S_{j'}^{L-1}(t) \right\rangle_{S_{j'}^{L-1}|L-2, j'}}{\partial w_{ij}^{L-2}} \approx \frac{w_{j'i}^{L-1}}{\left(\Delta u\right)^2} S_{j'}^{L-1}(t) \left( \epsilon \ast \left[ S_i^{L-2} \left( \epsilon \ast S_j^{L-3} \right) \right] \right)(t) \;.
\end{equation}
Substituting the above into Eq~\eqref{eq:grad_wh2_3} yields the gradient of a neuron's membrane potential in the final layer:
\begin{equation} \label{eq:grad_wh2_11}
\frac{\partial u_{i'}^L(t)}{\partial w_{ij}^{L-2}}\bigg\rvert_{t = \tau_{i'}} = \frac{1}{\left(\Delta u\right)^2} \sum_{j' \in \Gamma_{i'}^L} w_{i'j'}^L w_{j'i}^{L-1} \left( \epsilon \ast \left[ S_{j'}^{L-1} \left( \epsilon \ast \left[ S_i^{L-2} \left( \epsilon \ast S_j^{L-3} \right) \right] \right) \right] \right)(\tau_{i'}) \;,
\end{equation}
Combining the above with Eqs~\eqref{eq:grad_wh2} and \eqref{eq:grad_wh2_2} gives rise to the weight update rule for the third-last layer:
\begin{equation} \label{eq:dw_L-2}
\Delta w_{ij}^{L-2} = -\frac{\eta}{\left(\Delta u\right)^2} \sum_{i' \in I'} \delta_{i'}^L \sum_{j' \in \Gamma_{i'}^L} w_{i'j'}^L w_{j'i}^{L-1} \left( \epsilon \ast \left[ S_{j'}^{L-1} \left( \epsilon \ast \left[ S_i^{L-2} \left( \epsilon \ast S_j^{L-3} \right) \right] \right) \right] \right)(\tau_{i'}) \;.
\end{equation}
The above weight update formula depends on chained convolution operations, deriving from the intermediate synaptic processes occuring between distant, upstream spiking events and downstream target activity. By following a similar approach to the above steps, hidden weight updates for deeper network architectures can also be derived sharing this pattern.

\section*{Acknowledgments}

This research has received funding from the European Union’s Horizon 2020
Framework Programme for Research and Innovation under the Specific Grant Agreement
No. 785907 (Human Brain Project SGA2).

\bibliographystyle{apacite}

\begin{thebibliography}{}

\bibitem [\protect \citeauthoryear {%
Albers%
, Westkott%
\BCBL {}\ \BBA {} Pawelzik%
}{%
Albers%
\ \protect \BOthers {.}}{%
{\protect \APACyear {2016}}%
}]{%
Albers2016}
\APACinsertmetastar {%
Albers2016}%
\begin{APACrefauthors}%
Albers, C.%
, Westkott, M.%
\BCBL {}\ \BBA {} Pawelzik, K.%
\end{APACrefauthors}%
\unskip\
\newblock
\APACrefYearMonthDay{2016}{}{}.
\newblock
{\BBOQ}\APACrefatitle {Learning of precise spike times with homeostatic
  membrane potential dependent synaptic plasticity} {Learning of precise spike
  times with homeostatic membrane potential dependent synaptic
  plasticity}.{\BBCQ}
\newblock
\APACjournalVolNumPages{PLoS ONE}{11}{2}{}.
\PrintBackRefs{\CurrentBib}

\bibitem [\protect \citeauthoryear {%
Bagheri%
, Simeone%
\BCBL {}\ \BBA {} Rajendran%
}{%
Bagheri%
\ \protect \BOthers {.}}{%
{\protect \APACyear {2018}}%
}]{%
Bagheri2018}
\APACinsertmetastar {%
Bagheri2018}%
\begin{APACrefauthors}%
Bagheri, A.%
, Simeone, O.%
\BCBL {}\ \BBA {} Rajendran, B.%
\end{APACrefauthors}%
\unskip\
\newblock
\APACrefYearMonthDay{2018}{}{}.
\newblock
{\BBOQ}\APACrefatitle {Training probabilistic spiking neural networks with
  first-to-spike decoding} {Training probabilistic spiking neural networks with
  first-to-spike decoding}.{\BBCQ}
\newblock
\BIn{} \APACrefbtitle {{2018 IEEE International Conference on Acoustics, Speech
  and Signal Processing (ICASSP)}} {{2018 IEEE International Conference on
  Acoustics, Speech and Signal Processing (ICASSP)}}\ (\BPGS\ 2986--2990).
\PrintBackRefs{\CurrentBib}

\bibitem [\protect \citeauthoryear {%
Bi%
\ \BBA {} Poo%
}{%
Bi%
\ \BBA {} Poo%
}{%
{\protect \APACyear {1998}}%
}]{%
Bi1998}
\APACinsertmetastar {%
Bi1998}%
\begin{APACrefauthors}%
Bi, G\BHBI Q.%
\BCBT {}\ \BBA {} Poo, M\BHBI M.%
\end{APACrefauthors}%
\unskip\
\newblock
\APACrefYearMonthDay{1998}{}{}.
\newblock
{\BBOQ}\APACrefatitle {Synaptic modifications in cultured hippocampal neurons:
  dependence on spike timing, synaptic strength, and postsynaptic cell type}
  {Synaptic modifications in cultured hippocampal neurons: dependence on spike
  timing, synaptic strength, and postsynaptic cell type}.{\BBCQ}
\newblock
\APACjournalVolNumPages{Journal of Neuroscience}{18}{24}{10464--10472}.
\PrintBackRefs{\CurrentBib}

\bibitem [\protect \citeauthoryear {%
Bohte%
, Kok%
\BCBL {}\ \BBA {} Poutr{\'e}%
}{%
Bohte%
\ \protect \BOthers {.}}{%
{\protect \APACyear {2002}}%
}]{%
Bohte2002a}
\APACinsertmetastar {%
Bohte2002a}%
\begin{APACrefauthors}%
Bohte, S\BPBI M.%
, Kok, J\BPBI N.%
\BCBL {}\ \BBA {} Poutr{\'e}, H\BPBI L.%
\end{APACrefauthors}%
\unskip\
\newblock
\APACrefYearMonthDay{2002}{}{}.
\newblock
{\BBOQ}\APACrefatitle {Error-backpropagation in temporally encoded networks of
  spiking neurons} {Error-backpropagation in temporally encoded networks of
  spiking neurons}.{\BBCQ}
\newblock
\APACjournalVolNumPages{Neurocomputing}{48}{1}{17--37}.
\PrintBackRefs{\CurrentBib}

\bibitem [\protect \citeauthoryear {%
Booij%
\ \BBA {} Nguyen%
}{%
Booij%
\ \BBA {} Nguyen%
}{%
{\protect \APACyear {2005}}%
}]{%
Booij2005}
\APACinsertmetastar {%
Booij2005}%
\begin{APACrefauthors}%
Booij, O.%
\BCBT {}\ \BBA {} Nguyen, H\BPBI T.%
\end{APACrefauthors}%
\unskip\
\newblock
\APACrefYearMonthDay{2005}{}{}.
\newblock
{\BBOQ}\APACrefatitle {A gradient descent rule for spiking neurons emitting
  multiple spikes} {A gradient descent rule for spiking neurons emitting
  multiple spikes}.{\BBCQ}
\newblock
\APACjournalVolNumPages{Information Processing Letters}{95}{6}{552--558}.
\PrintBackRefs{\CurrentBib}

\bibitem [\protect \citeauthoryear {%
Brea%
, Senn%
\BCBL {}\ \BBA {} Pfister%
}{%
Brea%
\ \protect \BOthers {.}}{%
{\protect \APACyear {2013}}%
}]{%
Brea2013}
\APACinsertmetastar {%
Brea2013}%
\begin{APACrefauthors}%
Brea, J.%
, Senn, W.%
\BCBL {}\ \BBA {} Pfister, J\BHBI P.%
\end{APACrefauthors}%
\unskip\
\newblock
\APACrefYearMonthDay{2013}{}{}.
\newblock
{\BBOQ}\APACrefatitle {Matching recall and storage in sequence learning with
  spiking neural networks} {Matching recall and storage in sequence learning
  with spiking neural networks}.{\BBCQ}
\newblock
\APACjournalVolNumPages{Journal of Neuroscience}{33}{23}{9565--9575}.
\PrintBackRefs{\CurrentBib}

\bibitem [\protect \citeauthoryear {%
Deng%
\ \protect \BOthers {.}}{%
Deng%
\ \protect \BOthers {.}}{%
{\protect \APACyear {2009}}%
}]{%
Deng2009}
\APACinsertmetastar {%
Deng2009}%
\begin{APACrefauthors}%
Deng, J.%
, Dong, W.%
, Socher, R.%
, Li, L\BHBI J.%
, Li, K.%
\BCBL {}\ \BBA {} Fei-Fei, L.%
\end{APACrefauthors}%
\unskip\
\newblock
\APACrefYearMonthDay{2009}{}{}.
\newblock
{\BBOQ}\APACrefatitle {Imagenet: A large-scale hierarchical image database}
  {Imagenet: A large-scale hierarchical image database}.{\BBCQ}
\newblock
\BIn{} \APACrefbtitle {{2009 IEEE Conference on Computer Vision and Pattern
  Recognition}} {{2009 IEEE Conference on Computer Vision and Pattern
  Recognition}}\ (\BPGS\ 248--255).
\PrintBackRefs{\CurrentBib}

\bibitem [\protect \citeauthoryear {%
Diehl%
\ \protect \BOthers {.}}{%
Diehl%
\ \protect \BOthers {.}}{%
{\protect \APACyear {2015}}%
}]{%
Diehl2015}
\APACinsertmetastar {%
Diehl2015}%
\begin{APACrefauthors}%
Diehl, P\BPBI U.%
, Neil, D.%
, Binas, J.%
, Cook, M.%
, Liu, S\BHBI C.%
\BCBL {}\ \BBA {} Pfeiffer, M.%
\end{APACrefauthors}%
\unskip\
\newblock
\APACrefYearMonthDay{2015}{}{}.
\newblock
{\BBOQ}\APACrefatitle {Fast-classifying, high-accuracy spiking deep networks
  through weight and threshold balancing} {Fast-classifying, high-accuracy
  spiking deep networks through weight and threshold balancing}.{\BBCQ}
\newblock
\BIn{} \APACrefbtitle {{2015 International Joint Conference on Neural Networks
  (IJCNN)}} {{2015 International Joint Conference on Neural Networks (IJCNN)}}\
  (\BPGS\ 1--8).
\PrintBackRefs{\CurrentBib}

\bibitem [\protect \citeauthoryear {%
Fisher%
}{%
Fisher%
}{%
{\protect \APACyear {1936}}%
}]{%
Fisher1936}
\APACinsertmetastar {%
Fisher1936}%
\begin{APACrefauthors}%
Fisher, R\BPBI A.%
\end{APACrefauthors}%
\unskip\
\newblock
\APACrefYearMonthDay{1936}{}{}.
\newblock
{\BBOQ}\APACrefatitle {The use of multiple measurements in taxonomic problems}
  {The use of multiple measurements in taxonomic problems}.{\BBCQ}
\newblock
\APACjournalVolNumPages{Annals of Eugenics}{7}{2}{179--188}.
\PrintBackRefs{\CurrentBib}

\bibitem [\protect \citeauthoryear {%
Florian%
}{%
Florian%
}{%
{\protect \APACyear {2012}}%
}]{%
Florian2012}
\APACinsertmetastar {%
Florian2012}%
\begin{APACrefauthors}%
Florian, R\BPBI V.%
\end{APACrefauthors}%
\unskip\
\newblock
\APACrefYearMonthDay{2012}{}{}.
\newblock
{\BBOQ}\APACrefatitle {The Chronotron: A Neuron That Learns to Fire Temporally
  Precise Spike Patterns} {The chronotron: A neuron that learns to fire
  temporally precise spike patterns}.{\BBCQ}
\newblock
\APACjournalVolNumPages{PLoS ONE}{7}{8}{e40233}.
\PrintBackRefs{\CurrentBib}

\bibitem [\protect \citeauthoryear {%
Fr{\'e}maux%
, Sprekeler%
\BCBL {}\ \BBA {} Gerstner%
}{%
Fr{\'e}maux%
\ \protect \BOthers {.}}{%
{\protect \APACyear {2013}}%
}]{%
Fremaux2013}
\APACinsertmetastar {%
Fremaux2013}%
\begin{APACrefauthors}%
Fr{\'e}maux, N.%
, Sprekeler, H.%
\BCBL {}\ \BBA {} Gerstner, W.%
\end{APACrefauthors}%
\unskip\
\newblock
\APACrefYearMonthDay{2013}{}{}.
\newblock
{\BBOQ}\APACrefatitle {Reinforcement learning using a continuous time
  actor-critic framework with spiking neurons} {Reinforcement learning using a
  continuous time actor-critic framework with spiking neurons}.{\BBCQ}
\newblock
\APACjournalVolNumPages{PLoS Computational Biology}{9}{4}{e1003024}.
\PrintBackRefs{\CurrentBib}

\bibitem [\protect \citeauthoryear {%
Friedmann%
\ \protect \BOthers {.}}{%
Friedmann%
\ \protect \BOthers {.}}{%
{\protect \APACyear {2017}}%
}]{%
Friedmann2017}
\APACinsertmetastar {%
Friedmann2017}%
\begin{APACrefauthors}%
Friedmann, S.%
, Schemmel, J.%
, Gr{\"u}bl, A.%
, Hartel, A.%
, Hock, M.%
\BCBL {}\ \BBA {} Meier, K.%
\end{APACrefauthors}%
\unskip\
\newblock
\APACrefYearMonthDay{2017}{}{}.
\newblock
{\BBOQ}\APACrefatitle {Demonstrating hybrid learning in a flexible neuromorphic
  hardware system} {Demonstrating hybrid learning in a flexible neuromorphic
  hardware system}.{\BBCQ}
\newblock
\APACjournalVolNumPages{IEEE Transactions on Biomedical Circuits and
  Systems}{11}{1}{128--142}.
\PrintBackRefs{\CurrentBib}

\bibitem [\protect \citeauthoryear {%
Gardner%
\ \BBA {} Gr{\"u}ning%
}{%
Gardner%
\ \BBA {} Gr{\"u}ning%
}{%
{\protect \APACyear {2016}}%
}]{%
Gardner2016}
\APACinsertmetastar {%
Gardner2016}%
\begin{APACrefauthors}%
Gardner, B.%
\BCBT {}\ \BBA {} Gr{\"u}ning, A.%
\end{APACrefauthors}%
\unskip\
\newblock
\APACrefYearMonthDay{2016}{}{}.
\newblock
{\BBOQ}\APACrefatitle {Supervised Learning in Spiking Neural Networks for
  Precise Temporal Encoding} {Supervised learning in spiking neural networks
  for precise temporal encoding}.{\BBCQ}
\newblock
\APACjournalVolNumPages{PLoS ONE}{11}{8}{e0161335}.
\PrintBackRefs{\CurrentBib}

\bibitem [\protect \citeauthoryear {%
Gardner%
, Sporea%
\BCBL {}\ \BBA {} Gr{\"u}ning%
}{%
Gardner%
\ \protect \BOthers {.}}{%
{\protect \APACyear {2015}}%
}]{%
Gardner2015}
\APACinsertmetastar {%
Gardner2015}%
\begin{APACrefauthors}%
Gardner, B.%
, Sporea, I.%
\BCBL {}\ \BBA {} Gr{\"u}ning, A.%
\end{APACrefauthors}%
\unskip\
\newblock
\APACrefYearMonthDay{2015}{}{}.
\newblock
{\BBOQ}\APACrefatitle {Learning Spatiotemporally Encoded Pattern
  Transformations in Structured Spiking Neural Networks} {Learning
  spatiotemporally encoded pattern transformations in structured spiking neural
  networks}.{\BBCQ}
\newblock
\APACjournalVolNumPages{Neural Computation}{27}{12}{2548--2586}.
\PrintBackRefs{\CurrentBib}

\bibitem [\protect \citeauthoryear {%
Gerstner%
\ \BBA {} Kistler%
}{%
Gerstner%
\ \BBA {} Kistler%
}{%
{\protect \APACyear {2002}}%
}]{%
Gerstner2002}
\APACinsertmetastar {%
Gerstner2002}%
\begin{APACrefauthors}%
Gerstner, W.%
\BCBT {}\ \BBA {} Kistler, W\BPBI M.%
\end{APACrefauthors}%
\unskip\
\newblock
\APACrefYear{2002}.
\newblock
\APACrefbtitle {Spiking neuron models: Single neurons, populations, plasticity}
  {Spiking neuron models: Single neurons, populations, plasticity}.
\newblock
\APACaddressPublisher{}{Cambridge University Press}.
\PrintBackRefs{\CurrentBib}

\bibitem [\protect \citeauthoryear {%
Gerstner%
, Kistler%
, Naud%
\BCBL {}\ \BBA {} Paninski%
}{%
Gerstner%
\ \protect \BOthers {.}}{%
{\protect \APACyear {2014}}%
}]{%
Gerstner2014}
\APACinsertmetastar {%
Gerstner2014}%
\begin{APACrefauthors}%
Gerstner, W.%
, Kistler, W\BPBI M.%
, Naud, R.%
\BCBL {}\ \BBA {} Paninski, L.%
\end{APACrefauthors}%
\unskip\
\newblock
\APACrefYear{2014}.
\newblock
\APACrefbtitle {Neuronal dynamics: From single neurons to networks and models
  of cognition} {Neuronal dynamics: From single neurons to networks and models
  of cognition}.
\newblock
\APACaddressPublisher{}{Cambridge University Press}.
\PrintBackRefs{\CurrentBib}

\bibitem [\protect \citeauthoryear {%
Ghosh-Dastidar%
\ \BBA {} Adeli%
}{%
Ghosh-Dastidar%
\ \BBA {} Adeli%
}{%
{\protect \APACyear {2009}}%
}]{%
Ghosh2009}
\APACinsertmetastar {%
Ghosh2009}%
\begin{APACrefauthors}%
Ghosh-Dastidar, S.%
\BCBT {}\ \BBA {} Adeli, H.%
\end{APACrefauthors}%
\unskip\
\newblock
\APACrefYearMonthDay{2009}{}{}.
\newblock
{\BBOQ}\APACrefatitle {A new supervised learning algorithm for multiple spiking
  neural networks with application in epilepsy and seizure detection} {A new
  supervised learning algorithm for multiple spiking neural networks with
  application in epilepsy and seizure detection}.{\BBCQ}
\newblock
\APACjournalVolNumPages{Neural Networks}{22}{10}{1419--1431}.
\PrintBackRefs{\CurrentBib}

\bibitem [\protect \citeauthoryear {%
Gollisch%
\ \BBA {} Meister%
}{%
Gollisch%
\ \BBA {} Meister%
}{%
{\protect \APACyear {2008}}%
}]{%
Gollisch2008}
\APACinsertmetastar {%
Gollisch2008}%
\begin{APACrefauthors}%
Gollisch, T.%
\BCBT {}\ \BBA {} Meister, M.%
\end{APACrefauthors}%
\unskip\
\newblock
\APACrefYearMonthDay{2008}{}{}.
\newblock
{\BBOQ}\APACrefatitle {Rapid neural coding in the retina with relative spike
  latencies} {Rapid neural coding in the retina with relative spike
  latencies}.{\BBCQ}
\newblock
\APACjournalVolNumPages{Science}{319}{5866}{1108--1111}.
\PrintBackRefs{\CurrentBib}

\bibitem [\protect \citeauthoryear {%
Gr{\"u}ning%
\ \BBA {} Bohte%
}{%
Gr{\"u}ning%
\ \BBA {} Bohte%
}{%
{\protect \APACyear {2014}}%
}]{%
Gruning2014}
\APACinsertmetastar {%
Gruning2014}%
\begin{APACrefauthors}%
Gr{\"u}ning, A.%
\BCBT {}\ \BBA {} Bohte, S\BPBI M.%
\end{APACrefauthors}%
\unskip\
\newblock
\APACrefYearMonthDay{2014}{}{}.
\newblock
{\BBOQ}\APACrefatitle {Spiking Neural Networks: Principles and Challenges.}
  {Spiking neural networks: Principles and challenges.}{\BBCQ}
\newblock
\BIn{} \APACrefbtitle {{Proceedings of the 22nd European Symposium on
  Artificial Neural Networks (ESANN 2014)}.} {{Proceedings of the 22nd European
  Symposium on Artificial Neural Networks (ESANN 2014)}.}
\newblock
\APACaddressPublisher{}{Computational Intelligence and Machine Learning
  (Bruges: ESANN)}.
\PrintBackRefs{\CurrentBib}

\bibitem [\protect \citeauthoryear {%
Gr{\"u}ning%
\ \BBA {} Sporea%
}{%
Gr{\"u}ning%
\ \BBA {} Sporea%
}{%
{\protect \APACyear {2012}}%
}]{%
Gruning2012}
\APACinsertmetastar {%
Gruning2012}%
\begin{APACrefauthors}%
Gr{\"u}ning, A.%
\BCBT {}\ \BBA {} Sporea, I.%
\end{APACrefauthors}%
\unskip\
\newblock
\APACrefYearMonthDay{2012}{}{}.
\newblock
{\BBOQ}\APACrefatitle {Supervised learning of logical operations in layered
  spiking neural networks with spike train encoding} {Supervised learning of
  logical operations in layered spiking neural networks with spike train
  encoding}.{\BBCQ}
\newblock
\APACjournalVolNumPages{Neural Processing Letters}{36}{2}{117--134}.
\PrintBackRefs{\CurrentBib}

\bibitem [\protect \citeauthoryear {%
G{\"u}tig%
}{%
G{\"u}tig%
}{%
{\protect \APACyear {2014}}%
}]{%
Gutig2014}
\APACinsertmetastar {%
Gutig2014}%
\begin{APACrefauthors}%
G{\"u}tig, R.%
\end{APACrefauthors}%
\unskip\
\newblock
\APACrefYearMonthDay{2014}{}{}.
\newblock
{\BBOQ}\APACrefatitle {To spike, or when to spike?} {To spike, or when to
  spike?}{\BBCQ}
\newblock
\APACjournalVolNumPages{Current Opinion in Neurobiology}{25}{}{134--139}.
\PrintBackRefs{\CurrentBib}

\bibitem [\protect \citeauthoryear {%
G{\"u}tig%
\ \BBA {} Sompolinsky%
}{%
G{\"u}tig%
\ \BBA {} Sompolinsky%
}{%
{\protect \APACyear {2006}}%
}]{%
Gutig2006}
\APACinsertmetastar {%
Gutig2006}%
\begin{APACrefauthors}%
G{\"u}tig, R.%
\BCBT {}\ \BBA {} Sompolinsky, H.%
\end{APACrefauthors}%
\unskip\
\newblock
\APACrefYearMonthDay{2006}{}{}.
\newblock
{\BBOQ}\APACrefatitle {The tempotron: a neuron that learns spike timing--based
  decisions} {The tempotron: a neuron that learns spike timing--based
  decisions}.{\BBCQ}
\newblock
\APACjournalVolNumPages{Nature Neuroscience}{9}{3}{420--428}.
\PrintBackRefs{\CurrentBib}

\bibitem [\protect \citeauthoryear {%
G{\"u}tig%
\ \BBA {} Sompolinsky%
}{%
G{\"u}tig%
\ \BBA {} Sompolinsky%
}{%
{\protect \APACyear {2009}}%
}]{%
Gutig2009}
\APACinsertmetastar {%
Gutig2009}%
\begin{APACrefauthors}%
G{\"u}tig, R.%
\BCBT {}\ \BBA {} Sompolinsky, H.%
\end{APACrefauthors}%
\unskip\
\newblock
\APACrefYearMonthDay{2009}{}{}.
\newblock
{\BBOQ}\APACrefatitle {Time-warp--invariant neuronal processing}
  {Time-warp--invariant neuronal processing}.{\BBCQ}
\newblock
\APACjournalVolNumPages{PLoS Biology}{7}{7}{}.
\PrintBackRefs{\CurrentBib}

\bibitem [\protect \citeauthoryear {%
Hinton%
, Srivastava%
\BCBL {}\ \BBA {} Swersky%
}{%
Hinton%
\ \protect \BOthers {.}}{%
{\protect \APACyear {2012}}%
}]{%
Hinton2012}
\APACinsertmetastar {%
Hinton2012}%
\begin{APACrefauthors}%
Hinton, G.%
, Srivastava, N.%
\BCBL {}\ \BBA {} Swersky, K.%
\end{APACrefauthors}%
\unskip\
\newblock
\APACrefYearMonthDay{2012}{}{}.
\newblock
{\BBOQ}\APACrefatitle {Neural networks for machine learning} {Neural networks
  for machine learning}.{\BBCQ}
\newblock
\APACjournalVolNumPages{Coursera, video lectures}{}{}{}.
\PrintBackRefs{\CurrentBib}

\bibitem [\protect \citeauthoryear {%
Hung%
, Kreiman%
, Poggio%
\BCBL {}\ \BBA {} DiCarlo%
}{%
Hung%
\ \protect \BOthers {.}}{%
{\protect \APACyear {2005}}%
}]{%
Hung2005}
\APACinsertmetastar {%
Hung2005}%
\begin{APACrefauthors}%
Hung, C\BPBI P.%
, Kreiman, G.%
, Poggio, T.%
\BCBL {}\ \BBA {} DiCarlo, J\BPBI J.%
\end{APACrefauthors}%
\unskip\
\newblock
\APACrefYearMonthDay{2005}{}{}.
\newblock
{\BBOQ}\APACrefatitle {Fast readout of object identity from macaque inferior
  temporal cortex} {Fast readout of object identity from macaque inferior
  temporal cortex}.{\BBCQ}
\newblock
\APACjournalVolNumPages{Science}{310}{5749}{863--866}.
\PrintBackRefs{\CurrentBib}

\bibitem [\protect \citeauthoryear {%
Jang%
, Simeone%
, Gardner%
\BCBL {}\ \BBA {} Gruning%
}{%
Jang%
\ \protect \BOthers {.}}{%
{\protect \APACyear {2019}}%
}]{%
Jang2019}
\APACinsertmetastar {%
Jang2019}%
\begin{APACrefauthors}%
Jang, H.%
, Simeone, O.%
, Gardner, B.%
\BCBL {}\ \BBA {} Gruning, A.%
\end{APACrefauthors}%
\unskip\
\newblock
\APACrefYearMonthDay{2019}{}{}.
\newblock
{\BBOQ}\APACrefatitle {An Introduction to Probabilistic Spiking Neural
  Networks: Probabilistic Models, Learning Rules, and Applications} {An
  introduction to probabilistic spiking neural networks: Probabilistic models,
  learning rules, and applications}.{\BBCQ}
\newblock
\APACjournalVolNumPages{IEEE Signal Processing Magazine}{36}{6}{64--77}.
\PrintBackRefs{\CurrentBib}

\bibitem [\protect \citeauthoryear {%
Jang%
, Skatchkovsky%
\BCBL {}\ \BBA {} Simeone%
}{%
Jang%
\ \protect \BOthers {.}}{%
{\protect \APACyear {2020}}%
}]{%
Jang2020}
\APACinsertmetastar {%
Jang2020}%
\begin{APACrefauthors}%
Jang, H.%
, Skatchkovsky, N.%
\BCBL {}\ \BBA {} Simeone, O.%
\end{APACrefauthors}%
\unskip\
\newblock
\APACrefYearMonthDay{2020}{}{}.
\newblock
{\BBOQ}\APACrefatitle {VOWEL: A Local Online Learning Rule for Recurrent
  Networks of Probabilistic Spiking Winner-Take-All Circuits} {Vowel: A local
  online learning rule for recurrent networks of probabilistic spiking
  winner-take-all circuits}.{\BBCQ}
\newblock
\APACjournalVolNumPages{arXiv preprint arXiv:2004.09416}{}{}{}.
\PrintBackRefs{\CurrentBib}

\bibitem [\protect \citeauthoryear {%
Jimenez~Rezende%
\ \BBA {} Gerstner%
}{%
Jimenez~Rezende%
\ \BBA {} Gerstner%
}{%
{\protect \APACyear {2014}}%
}]{%
Rezende2014}
\APACinsertmetastar {%
Rezende2014}%
\begin{APACrefauthors}%
Jimenez~Rezende, D.%
\BCBT {}\ \BBA {} Gerstner, W.%
\end{APACrefauthors}%
\unskip\
\newblock
\APACrefYearMonthDay{2014}{}{}.
\newblock
{\BBOQ}\APACrefatitle {Stochastic variational learning in recurrent spiking
  networks} {Stochastic variational learning in recurrent spiking
  networks}.{\BBCQ}
\newblock
\APACjournalVolNumPages{Frontiers in Computational Neuroscience}{8}{}{38}.
\PrintBackRefs{\CurrentBib}

\bibitem [\protect \citeauthoryear {%
Kiani%
, Esteky%
\BCBL {}\ \BBA {} Tanaka%
}{%
Kiani%
\ \protect \BOthers {.}}{%
{\protect \APACyear {2005}}%
}]{%
Kiani2005}
\APACinsertmetastar {%
Kiani2005}%
\begin{APACrefauthors}%
Kiani, R.%
, Esteky, H.%
\BCBL {}\ \BBA {} Tanaka, K.%
\end{APACrefauthors}%
\unskip\
\newblock
\APACrefYearMonthDay{2005}{}{}.
\newblock
{\BBOQ}\APACrefatitle {Differences in onset latency of macaque inferotemporal
  neural responses to primate and non-primate faces} {Differences in onset
  latency of macaque inferotemporal neural responses to primate and non-primate
  faces}.{\BBCQ}
\newblock
\APACjournalVolNumPages{Journal of Neurophysiology}{94}{2}{1587--1596}.
\PrintBackRefs{\CurrentBib}

\bibitem [\protect \citeauthoryear {%
LeCun%
, Bottou%
, Bengio%
\BCBL {}\ \BBA {} Haffner%
}{%
LeCun%
\ \protect \BOthers {.}}{%
{\protect \APACyear {1998}}%
}]{%
LeCun1998}
\APACinsertmetastar {%
LeCun1998}%
\begin{APACrefauthors}%
LeCun, Y.%
, Bottou, L.%
, Bengio, Y.%
\BCBL {}\ \BBA {} Haffner, P.%
\end{APACrefauthors}%
\unskip\
\newblock
\APACrefYearMonthDay{1998}{}{}.
\newblock
{\BBOQ}\APACrefatitle {Gradient-based learning applied to document recognition}
  {Gradient-based learning applied to document recognition}.{\BBCQ}
\newblock
\APACjournalVolNumPages{Proceedings of the IEEE}{86}{11}{2278--2324}.
\PrintBackRefs{\CurrentBib}

\bibitem [\protect \citeauthoryear {%
Lee%
, Delbruck%
\BCBL {}\ \BBA {} Pfeiffer%
}{%
Lee%
\ \protect \BOthers {.}}{%
{\protect \APACyear {2016}}%
}]{%
Lee2016}
\APACinsertmetastar {%
Lee2016}%
\begin{APACrefauthors}%
Lee, J\BPBI H.%
, Delbruck, T.%
\BCBL {}\ \BBA {} Pfeiffer, M.%
\end{APACrefauthors}%
\unskip\
\newblock
\APACrefYearMonthDay{2016}{}{}.
\newblock
{\BBOQ}\APACrefatitle {Training deep spiking neural networks using
  backpropagation} {Training deep spiking neural networks using
  backpropagation}.{\BBCQ}
\newblock
\APACjournalVolNumPages{Frontiers in Neuroscience}{10}{}{508}.
\PrintBackRefs{\CurrentBib}

\bibitem [\protect \citeauthoryear {%
Lin%
\ \protect \BOthers {.}}{%
Lin%
\ \protect \BOthers {.}}{%
{\protect \APACyear {2018}}%
}]{%
Lin2018}
\APACinsertmetastar {%
Lin2018}%
\begin{APACrefauthors}%
Lin, C\BHBI K.%
, Wild, A.%
, Chinya, G\BPBI N.%
, Cao, Y.%
, Davies, M.%
, Lavery, D\BPBI M.%
\BCBL {}\ \BBA {} Wang, H.%
\end{APACrefauthors}%
\unskip\
\newblock
\APACrefYearMonthDay{2018}{}{}.
\newblock
{\BBOQ}\APACrefatitle {Programming Spiking Neural Networks on Intel’s Loihi}
  {Programming spiking neural networks on intel’s loihi}.{\BBCQ}
\newblock
\APACjournalVolNumPages{Computer}{51}{3}{52--61}.
\PrintBackRefs{\CurrentBib}

\bibitem [\protect \citeauthoryear {%
Memmesheimer%
, Rubin%
, {\"O}lveczky%
\BCBL {}\ \BBA {} Sompolinsky%
}{%
Memmesheimer%
\ \protect \BOthers {.}}{%
{\protect \APACyear {2014}}%
}]{%
Memmesheimer2014}
\APACinsertmetastar {%
Memmesheimer2014}%
\begin{APACrefauthors}%
Memmesheimer, R\BHBI M.%
, Rubin, R.%
, {\"O}lveczky, B\BPBI P.%
\BCBL {}\ \BBA {} Sompolinsky, H.%
\end{APACrefauthors}%
\unskip\
\newblock
\APACrefYearMonthDay{2014}{}{}.
\newblock
{\BBOQ}\APACrefatitle {Learning precisely timed spikes} {Learning precisely
  timed spikes}.{\BBCQ}
\newblock
\APACjournalVolNumPages{Neuron}{82}{4}{925--938}.
\PrintBackRefs{\CurrentBib}

\bibitem [\protect \citeauthoryear {%
Mohemmed%
, Schliebs%
, Matsuda%
\BCBL {}\ \BBA {} Kasabov%
}{%
Mohemmed%
\ \protect \BOthers {.}}{%
{\protect \APACyear {2012}}%
}]{%
Mohemmed2012}
\APACinsertmetastar {%
Mohemmed2012}%
\begin{APACrefauthors}%
Mohemmed, A.%
, Schliebs, S.%
, Matsuda, S.%
\BCBL {}\ \BBA {} Kasabov, N.%
\end{APACrefauthors}%
\unskip\
\newblock
\APACrefYearMonthDay{2012}{}{}.
\newblock
{\BBOQ}\APACrefatitle {Span: Spike pattern association neuron for learning
  spatio-temporal spike patterns} {Span: Spike pattern association neuron for
  learning spatio-temporal spike patterns}.{\BBCQ}
\newblock
\APACjournalVolNumPages{International Journal of Neural
  Systems}{22}{04}{1250012}.
\PrintBackRefs{\CurrentBib}

\bibitem [\protect \citeauthoryear {%
Morrison%
, Diesmann%
\BCBL {}\ \BBA {} Gerstner%
}{%
Morrison%
\ \protect \BOthers {.}}{%
{\protect \APACyear {2008}}%
}]{%
Morrison2008}
\APACinsertmetastar {%
Morrison2008}%
\begin{APACrefauthors}%
Morrison, A.%
, Diesmann, M.%
\BCBL {}\ \BBA {} Gerstner, W.%
\end{APACrefauthors}%
\unskip\
\newblock
\APACrefYearMonthDay{2008}{}{}.
\newblock
{\BBOQ}\APACrefatitle {Phenomenological models of synaptic plasticity based on
  spike timing} {Phenomenological models of synaptic plasticity based on spike
  timing}.{\BBCQ}
\newblock
\APACjournalVolNumPages{Biological Cybernetics}{98}{6}{459--478}.
\PrintBackRefs{\CurrentBib}

\bibitem [\protect \citeauthoryear {%
Mostafa%
}{%
Mostafa%
}{%
{\protect \APACyear {2017}}%
}]{%
Mostafa2017}
\APACinsertmetastar {%
Mostafa2017}%
\begin{APACrefauthors}%
Mostafa, H.%
\end{APACrefauthors}%
\unskip\
\newblock
\APACrefYearMonthDay{2017}{}{}.
\newblock
{\BBOQ}\APACrefatitle {Supervised learning based on temporal coding in spiking
  neural networks} {Supervised learning based on temporal coding in spiking
  neural networks}.{\BBCQ}
\newblock
\APACjournalVolNumPages{IEEE Transactions on Neural Networks and Learning
  Systems}{}{}{}.
\PrintBackRefs{\CurrentBib}

\bibitem [\protect \citeauthoryear {%
E.~Neftci%
, Das%
, Pedroni%
, Kreutz-Delgado%
\BCBL {}\ \BBA {} Cauwenberghs%
}{%
E.~Neftci%
\ \protect \BOthers {.}}{%
{\protect \APACyear {2014}}%
}]{%
Neftci2014}
\APACinsertmetastar {%
Neftci2014}%
\begin{APACrefauthors}%
Neftci, E.%
, Das, S.%
, Pedroni, B.%
, Kreutz-Delgado, K.%
\BCBL {}\ \BBA {} Cauwenberghs, G.%
\end{APACrefauthors}%
\unskip\
\newblock
\APACrefYearMonthDay{2014}{}{}.
\newblock
{\BBOQ}\APACrefatitle {Event-driven contrastive divergence for spiking
  neuromorphic systems} {Event-driven contrastive divergence for spiking
  neuromorphic systems}.{\BBCQ}
\newblock
\APACjournalVolNumPages{Frontiers in Neuroscience}{7}{}{272}.
\PrintBackRefs{\CurrentBib}

\bibitem [\protect \citeauthoryear {%
E\BPBI O.~Neftci%
, Mostafa%
\BCBL {}\ \BBA {} Zenke%
}{%
E\BPBI O.~Neftci%
\ \protect \BOthers {.}}{%
{\protect \APACyear {2019}}%
}]{%
Neftci2019}
\APACinsertmetastar {%
Neftci2019}%
\begin{APACrefauthors}%
Neftci, E\BPBI O.%
, Mostafa, H.%
\BCBL {}\ \BBA {} Zenke, F.%
\end{APACrefauthors}%
\unskip\
\newblock
\APACrefYearMonthDay{2019}{}{}.
\newblock
{\BBOQ}\APACrefatitle {Surrogate gradient learning in spiking neural networks:
  Bringing the power of gradient-based optimization to spiking neural networks}
  {Surrogate gradient learning in spiking neural networks: Bringing the power
  of gradient-based optimization to spiking neural networks}.{\BBCQ}
\newblock
\APACjournalVolNumPages{IEEE Signal Processing Magazine}{36}{6}{51--63}.
\PrintBackRefs{\CurrentBib}

\bibitem [\protect \citeauthoryear {%
O'Connor%
, Neil%
, Liu%
, Delbruck%
\BCBL {}\ \BBA {} Pfeiffer%
}{%
O'Connor%
\ \protect \BOthers {.}}{%
{\protect \APACyear {2013}}%
}]{%
Connor2013}
\APACinsertmetastar {%
Connor2013}%
\begin{APACrefauthors}%
O'Connor, P.%
, Neil, D.%
, Liu, S\BHBI C.%
, Delbruck, T.%
\BCBL {}\ \BBA {} Pfeiffer, M.%
\end{APACrefauthors}%
\unskip\
\newblock
\APACrefYearMonthDay{2013}{}{}.
\newblock
{\BBOQ}\APACrefatitle {Real-time classification and sensor fusion with a
  spiking deep belief network} {Real-time classification and sensor fusion with
  a spiking deep belief network}.{\BBCQ}
\newblock
\APACjournalVolNumPages{Frontiers in Neuroscience}{7}{}{178}.
\PrintBackRefs{\CurrentBib}

\bibitem [\protect \citeauthoryear {%
Pfister%
, Toyoizumi%
, Barber%
\BCBL {}\ \BBA {} Gerstner%
}{%
Pfister%
\ \protect \BOthers {.}}{%
{\protect \APACyear {2006}}%
}]{%
Pfister2006}
\APACinsertmetastar {%
Pfister2006}%
\begin{APACrefauthors}%
Pfister, J\BHBI P.%
, Toyoizumi, T.%
, Barber, D.%
\BCBL {}\ \BBA {} Gerstner, W.%
\end{APACrefauthors}%
\unskip\
\newblock
\APACrefYearMonthDay{2006}{}{}.
\newblock
{\BBOQ}\APACrefatitle {Optimal spike-timing-dependent plasticity for precise
  action potential firing in supervised learning} {Optimal
  spike-timing-dependent plasticity for precise action potential firing in
  supervised learning}.{\BBCQ}
\newblock
\APACjournalVolNumPages{Neural Computation}{18}{6}{1318--1348}.
\PrintBackRefs{\CurrentBib}

\bibitem [\protect \citeauthoryear {%
Ponulak%
\ \BBA {} Kasi{\'n}ski%
}{%
Ponulak%
\ \BBA {} Kasi{\'n}ski%
}{%
{\protect \APACyear {2010}}%
}]{%
Ponulak2010}
\APACinsertmetastar {%
Ponulak2010}%
\begin{APACrefauthors}%
Ponulak, F.%
\BCBT {}\ \BBA {} Kasi{\'n}ski, A.%
\end{APACrefauthors}%
\unskip\
\newblock
\APACrefYearMonthDay{2010}{}{}.
\newblock
{\BBOQ}\APACrefatitle {Supervised learning in spiking neural networks with
  ReSuMe: sequence learning, classification, and spike shifting} {Supervised
  learning in spiking neural networks with resume: sequence learning,
  classification, and spike shifting}.{\BBCQ}
\newblock
\APACjournalVolNumPages{Neural Computation}{22}{2}{467--510}.
\PrintBackRefs{\CurrentBib}

\bibitem [\protect \citeauthoryear {%
Sporea%
\ \BBA {} Gr{\"u}ning%
}{%
Sporea%
\ \BBA {} Gr{\"u}ning%
}{%
{\protect \APACyear {2013}}%
}]{%
Sporea2013}
\APACinsertmetastar {%
Sporea2013}%
\begin{APACrefauthors}%
Sporea, I.%
\BCBT {}\ \BBA {} Gr{\"u}ning, A.%
\end{APACrefauthors}%
\unskip\
\newblock
\APACrefYearMonthDay{2013}{}{}.
\newblock
{\BBOQ}\APACrefatitle {Supervised learning in multilayer spiking neural
  networks} {Supervised learning in multilayer spiking neural networks}.{\BBCQ}
\newblock
\APACjournalVolNumPages{Neural Computation}{25}{2}{473--509}.
\PrintBackRefs{\CurrentBib}

\bibitem [\protect \citeauthoryear {%
Tavanaei%
\ \BBA {} Maida%
}{%
Tavanaei%
\ \BBA {} Maida%
}{%
{\protect \APACyear {2019}}%
}]{%
Tavanaei2019}
\APACinsertmetastar {%
Tavanaei2019}%
\begin{APACrefauthors}%
Tavanaei, A.%
\BCBT {}\ \BBA {} Maida, A.%
\end{APACrefauthors}%
\unskip\
\newblock
\APACrefYearMonthDay{2019}{}{}.
\newblock
{\BBOQ}\APACrefatitle {BP-STDP: Approximating backpropagation using spike
  timing dependent plasticity} {Bp-stdp: Approximating backpropagation using
  spike timing dependent plasticity}.{\BBCQ}
\newblock
\APACjournalVolNumPages{Neurocomputing}{330}{}{39--47}.
\PrintBackRefs{\CurrentBib}

\bibitem [\protect \citeauthoryear {%
Urbanczik%
\ \BBA {} Senn%
}{%
Urbanczik%
\ \BBA {} Senn%
}{%
{\protect \APACyear {2009}}%
}]{%
Urbanczik2009}
\APACinsertmetastar {%
Urbanczik2009}%
\begin{APACrefauthors}%
Urbanczik, R.%
\BCBT {}\ \BBA {} Senn, W.%
\end{APACrefauthors}%
\unskip\
\newblock
\APACrefYearMonthDay{2009}{}{}.
\newblock
{\BBOQ}\APACrefatitle {A gradient learning rule for the tempotron} {A gradient
  learning rule for the tempotron}.{\BBCQ}
\newblock
\APACjournalVolNumPages{Neural Computation}{21}{2}{340--352}.
\PrintBackRefs{\CurrentBib}

\bibitem [\protect \citeauthoryear {%
van Rossum%
, Bi%
\BCBL {}\ \BBA {} Turrigiano%
}{%
van Rossum%
\ \protect \BOthers {.}}{%
{\protect \APACyear {2000}}%
}]{%
vanRossum2000}
\APACinsertmetastar {%
vanRossum2000}%
\begin{APACrefauthors}%
van Rossum, M\BPBI C.%
, Bi, G\BPBI Q.%
\BCBL {}\ \BBA {} Turrigiano, G\BPBI G.%
\end{APACrefauthors}%
\unskip\
\newblock
\APACrefYearMonthDay{2000}{}{}.
\newblock
{\BBOQ}\APACrefatitle {Stable Hebbian learning from spike timing-dependent
  plasticity} {Stable hebbian learning from spike timing-dependent
  plasticity}.{\BBCQ}
\newblock
\APACjournalVolNumPages{Journal of Neuroscience}{20}{23}{8812--8821}.
\PrintBackRefs{\CurrentBib}

\bibitem [\protect \citeauthoryear {%
Wolberg%
\ \BBA {} Mangasarian%
}{%
Wolberg%
\ \BBA {} Mangasarian%
}{%
{\protect \APACyear {1990}}%
}]{%
Wolberg1990}
\APACinsertmetastar {%
Wolberg1990}%
\begin{APACrefauthors}%
Wolberg, W\BPBI H.%
\BCBT {}\ \BBA {} Mangasarian, O\BPBI L.%
\end{APACrefauthors}%
\unskip\
\newblock
\APACrefYearMonthDay{1990}{}{}.
\newblock
{\BBOQ}\APACrefatitle {Multisurface method of pattern separation for medical
  diagnosis applied to breast cytology.} {Multisurface method of pattern
  separation for medical diagnosis applied to breast cytology.}{\BBCQ}
\newblock
\APACjournalVolNumPages{Proceedings of the National Academy of
  Sciences}{87}{23}{9193--9196}.
\PrintBackRefs{\CurrentBib}

\bibitem [\protect \citeauthoryear {%
Xiao%
, Rasul%
\BCBL {}\ \BBA {} Vollgraf%
}{%
Xiao%
\ \protect \BOthers {.}}{%
{\protect \APACyear {2017}}%
}]{%
Xiao2017}
\APACinsertmetastar {%
Xiao2017}%
\begin{APACrefauthors}%
Xiao, H.%
, Rasul, K.%
\BCBL {}\ \BBA {} Vollgraf, R.%
\end{APACrefauthors}%
\unskip\
\newblock
\APACrefYearMonthDay{2017}{}{}.
\newblock
{\BBOQ}\APACrefatitle {Fashion-mnist: a novel image dataset for benchmarking
  machine learning algorithms} {Fashion-mnist: a novel image dataset for
  benchmarking machine learning algorithms}.{\BBCQ}
\newblock
\APACjournalVolNumPages{arXiv preprint arXiv:1708.07747}{}{}{}.
\PrintBackRefs{\CurrentBib}

\bibitem [\protect \citeauthoryear {%
Yu%
, Tang%
, Tan%
\BCBL {}\ \BBA {} Li%
}{%
Yu%
\ \protect \BOthers {.}}{%
{\protect \APACyear {2013}}%
}]{%
Yu2013}
\APACinsertmetastar {%
Yu2013}%
\begin{APACrefauthors}%
Yu, Q.%
, Tang, H.%
, Tan, K\BPBI C.%
\BCBL {}\ \BBA {} Li, H.%
\end{APACrefauthors}%
\unskip\
\newblock
\APACrefYearMonthDay{2013}{}{}.
\newblock
{\BBOQ}\APACrefatitle {Precise-spike-driven synaptic plasticity: Learning
  hetero-association of spatiotemporal spike patterns} {Precise-spike-driven
  synaptic plasticity: Learning hetero-association of spatiotemporal spike
  patterns}.{\BBCQ}
\newblock
\APACjournalVolNumPages{PLoS ONE}{8}{11}{}.
\PrintBackRefs{\CurrentBib}

\bibitem [\protect \citeauthoryear {%
Zenke%
\ \BBA {} Ganguli%
}{%
Zenke%
\ \BBA {} Ganguli%
}{%
{\protect \APACyear {2018}}%
}]{%
Zenke2018}
\APACinsertmetastar {%
Zenke2018}%
\begin{APACrefauthors}%
Zenke, F.%
\BCBT {}\ \BBA {} Ganguli, S.%
\end{APACrefauthors}%
\unskip\
\newblock
\APACrefYearMonthDay{2018}{}{}.
\newblock
{\BBOQ}\APACrefatitle {Superspike: Supervised learning in multilayer spiking
  neural networks} {Superspike: Supervised learning in multilayer spiking
  neural networks}.{\BBCQ}
\newblock
\APACjournalVolNumPages{Neural Computation}{30}{6}{1514--1541}.
\PrintBackRefs{\CurrentBib}

\bibitem [\protect \citeauthoryear {%
Zenke%
\ \BBA {} Vogels%
}{%
Zenke%
\ \BBA {} Vogels%
}{%
{\protect \APACyear {2020}}%
}]{%
Zenke2020}
\APACinsertmetastar {%
Zenke2020}%
\begin{APACrefauthors}%
Zenke, F.%
\BCBT {}\ \BBA {} Vogels, T\BPBI P.%
\end{APACrefauthors}%
\unskip\
\newblock
\APACrefYearMonthDay{2020}{}{}.
\newblock
{\BBOQ}\APACrefatitle {The remarkable robustness of surrogate gradient learning
  for instilling complex function in spiking neural networks} {The remarkable
  robustness of surrogate gradient learning for instilling complex function in
  spiking neural networks}.{\BBCQ}
\newblock
\APACjournalVolNumPages{bioRxiv preprint 2020.06.29.176925}{}{}{}.
\PrintBackRefs{\CurrentBib}

\end{thebibliography}

\end{document}